\documentclass{article}

\PassOptionsToPackage{numbers, compress}{natbib}
\usepackage[final]{nips_2017}

\usepackage[utf8]{inputenc} 
\usepackage[T1]{fontenc}    
\usepackage{hyperref}       
\usepackage{url}            
\usepackage{booktabs}       
\usepackage{amsfonts}       
\usepackage{nicefrac}       
\usepackage{microtype}      
\usepackage{amsmath,amssymb,euscript,yfonts,psfrag,latexsym,graphicx}
\usepackage{bbm,color,amstext,wasysym,subfig,flushend,parskip,balance}
\graphicspath{{./},{./figures/}}

\newtheorem{thm}{Theorem}

\newcommand{\mR}{{\mathbb R}}		

\title{Affine Differential Invariants for\break Invariant Feature Point Detection}

\author{
  Stanley L. Tuznik\\
  Department of Applied Mathematics and Statistics\\
  Stony Brook University\\
  Stony Brook, NY 11794 \\
  \texttt{stanley.tuznik@stonybrook.edu} \\
  \And
  Peter J. Olver \\
  Department of Mathematics\\
  University of Minnesota\\
  Minneapolis, MN 55455\\
  \texttt{olver@math.umn.edu}\\
 \And
  Allen Tannenbaum \\
 Departments of Computer Science and Applied Mathematics/Statistics\\
 Stony Brook University\\
 Stony Brook, NY 11794\\
 \texttt{arobertan@gmail.com}\\
}

\begin{document}

\maketitle

\begin{abstract}
Image feature points are detected as pixels which locally maximize a detector function, two commonly used examples of which are the (Euclidean) image gradient and the Harris-Stephens corner detector. A major limitation of these feature detectors are that they are only Euclidean-invariant. In this work we demonstrate the application of a 2D affine-invariant image feature point detector based on differential invariants as derived through the equivariant method of moving frames. The fundamental equi-affine differential invariants for 3D image volumes are also computed.
\end{abstract}

\section{Introduction}
Image feature points have been widely used for decades as fundamental components of computer vision algorithms, and they continue to find use in such applications as medical image registration, estimation of homographies in stereo vision, and stitching of images for panoramas or satellite imagery. Features are generally defined as pixels (or sets of pixels) in an image which can be reliably identified and distinctly described across images. The goal of feature detection and description is to abstract images as point sets which can be matched to each other. These correspondences can then be used to compute (estimate) the transformation between the point sets and, thus, between the images.

Feature detectors are scalar functions which measure some sort of response on an image, and features are typically identified as local extrema in these functions. For example, the popular Harris-Stephens corner detector identifies feature points as pixels which locally maximize the function
\begin{equation}
R\left(x,y\right) = \hbox{det}\left(H\right)-\alpha \,\hbox{tr}\left(H\right)^2
\end{equation}
where $\alpha$ is a parameter and $H$ is the second moment matrix of the image \cite{har1988}. This function responds strongly on pixels where the image varies greatly in two directions, and thus is a corner detector. The popular SIFT feature algorithm uses a multi-scale image representation and computes a difference-of-Gaussians function (an approximation of the Laplacian-of-Gaussian) as a detector \cite{low2004}. SIFT features are local maxima of this function over both space and scale. This multi-scale representation allows identification of features at different spatial scales, and also serves to smooth noise out of the image.

\begin{figure}[t]
\centering
\subfloat[]{\includegraphics[scale=.1]{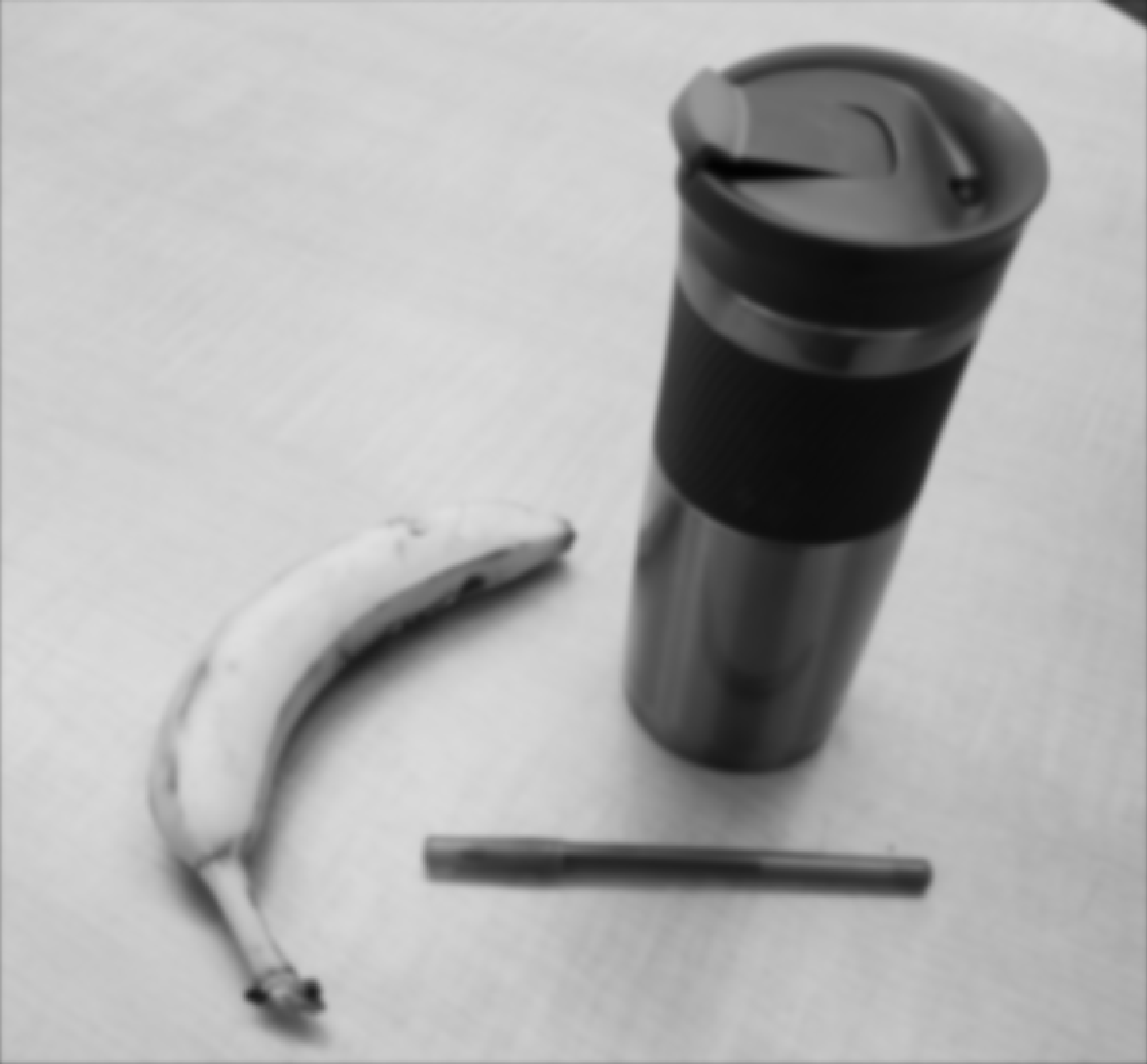}} 
\subfloat[]{\includegraphics[scale=.1]{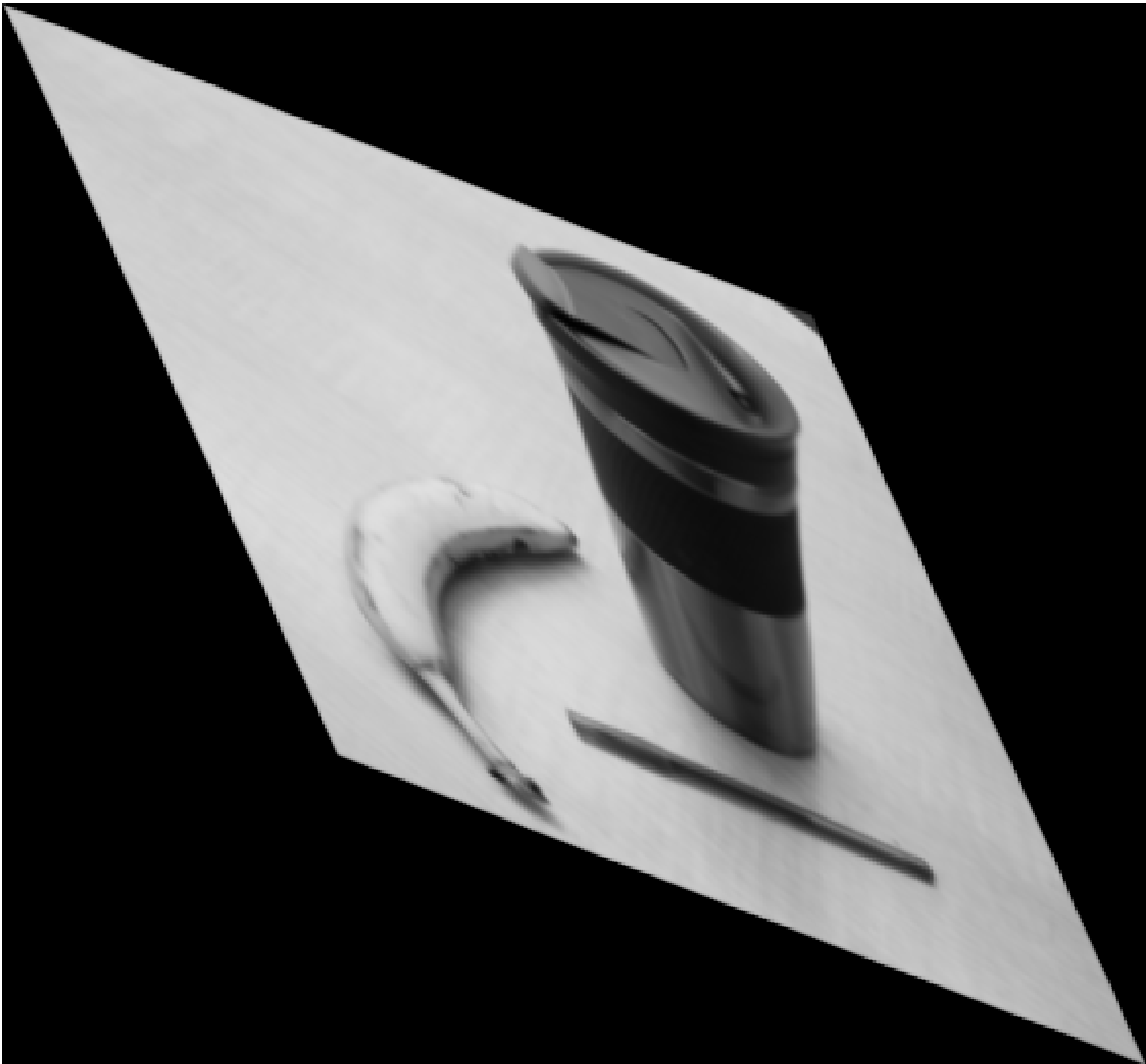}}
 \caption{Two images of a scene related by an equi-affine transformation.}
 \label{fig:ims}
\end{figure}

A major drawback to these detectors is that they are only Euclidean-invariant. In other words, these detectors will identify roughly the same image points as feature points in two images only if the two images are related by a Euclidean transformation (rotation, translation, reflection). The Euclidean group can generalized to the affine group by including stretching and skewing transformations (see Figure~\ref{fig:ims}). The equi-affine group, in particular, is useful in that is is a good approximation to projective transformations which are near the identity. Additionally, these transformations do not suffer from known difficulties associated with projective equivalence~\cite{ast1995}. Several affine-invariant image detectors have been proposed in the literature \cite{mik2005a,mik2005b}. In contrast, projective invariance and moving frame-based signatures have been successfully applied in vision~\cite{han2001,han2002}.

In this work, we propose the use of an invariant-theory based approach to affine-invariant feature detection. The equivariant method of moving frames \cite{olv2003,olv2015} was inspired by the classical moving frame theory, most closely associated with \'Elie Cartan --- see \cite{Gug} --- but whose history can be traced back to the early nineteenth century, \cite{AR}.  The equivariant method can be readily employed to compute the (differential) invariants of a given transformation (Lie) group. These invariants are then combined into an affine-invariant feature detector function \cite{olv1999}. Standard methods for feature description can then be applied to characterize each point. In contrast to the linear smoothing performed during the aforementioned multi-scale feature algorithms, we apply a nonlinear, affine-invariant scale-space to smooth our image.

The paper will have a tutorial flavor so as to remain fairly self-contained. In Section~\ref{sec:mf} we introduce the fundamental concepts of the method of moving frames and discuss how the method leads to the computation of differential invariants for a given Lie group acting on a manifold. In Section~\ref{sec:2d} the method is explicitly applied to the special affine group acting on coordinates of functions on $\mR^2$ (i.e., images), and the fundamental second-order differential invariants of this action are computed. The differential invariants for the equi-affine group acting on functions in $\mR^3$ are also computed. In Section~\ref{sec:fd} these invariants are used to compute a feature detector which is demonstrated to perform well at affine registration of 2D image pairs.

\section{Method of Moving Frames}
\label{sec:mf}
The method of moving frames is a powerful tool used to study the geometric properties of submanifolds under the action of transformation (Lie) groups \cite{olv2003,olv2015}. In particular, the method allows direct computation of the differential invariants of a transformation group. The equivarant method of moving frames was introduced in~\cite{fels1999} and permits a systematic application to general transformation groups acting on manifolds. The salient points of this theory are summarized here.

Given an $r$-dimensional Lie group, $G$, acting on an $m$-dimensional manifold, $M$, the method regards the construction of a moving frame, $\rho$. A moving frame $\rho$ is defined as a $G$-equivariant map $\rho:G\rightarrow M$. We will exclusively use right equivariant moving frames, which means that for $g\in G$ and $z$ a local coordinate on $M$, we have
$$ \rho\left(g\cdot z\right)= \rho \left(z\right) \cdot g^{-1} .$$
As we will demonstrate, the computation of a moving frame for a Lie group action will allow us to compute the fundamental differential invariants of the group action on the manifold.

A major theorem regarding the existence of such moving frames is as follows~\cite{olv2003}:
\begin{thm}
A moving frame exists near a point $z\in M$ if and only if $G$ acts freely and regularly near $z$.
\label{th:ex}
\end{thm}
The freeness condition is best expressed in terms of isotropy subgroups of $G$. The isotropy subgroup of $G$ with respect to $z\in M$ is the set
$$ G_z = \left\{ g ~|~ g\cdot z = z \right\}, $$
that is, the set of all group elements which fix $z$. The group action is free if
\begin{equation}
G_z = \left\{ e \right\} ~ \hbox{for all } z\in M
\end{equation}
where $e$ is the group identity. In other words, the action is free if for any $z\in M$, the only group element fixing $z$ is the identity. This requires that the orbits of $G$ have the same dimension as $G$ itself, and hence a necessary (but not sufficient) condition for freeness is that $\dim G \leq \dim M$.  The group action is regular if the orbits form a regular foliation of $M$.

If the group action is both free and regular, Cartan's method of normalization is used to compute a moving frame. We begin by choosing a coordinate cross-section to the group orbits. A cross section is a restriction of the form
\begin{equation}
K = \left\{ z_1=c_1, \ldots , z_r=c_r \right\},
\label{eq:cse}
\end{equation}
that is, we restrict $r$ of the coordinates to be fixed. This cross-section must be transverse to the group orbits. By freeness, for any $z\in M$, there is a group element $g$ mapping $z$ to this cross-section, i.e., to the unique point in the intersection of the cross-section with the group orbit through $z$. This unique $g$ is then the right moving frame for the action. This is formalized in the following theorem, which provides a method for practical construction of the right moving frame for a free and regular group action given a cross-section.
\begin{thm}
Let $G$ act freely and regularly on $M$, and let $K$ be a cross-section. For $z\in M$, let $g=\rho\left(z\right)$ be the unique group element mapping $z$ to this cross-section:
\begin{equation}
g\cdot z = \rho\left(z\right)\cdot z \in K 
\label{eq:norm}
\end{equation}
Then $\rho:M\rightarrow G$ is a right moving frame for the group action.
\end{thm}
The goal is to obtain the group transformation parameters $g = \left(g_1,g_2,\ldots,g_r\right)$ in equation~\eqref{eq:norm} by applying the cross-section~\eqref{eq:cse}. Explicitly, if we write 
\begin{equation}
w\left(g,z\right) = g\cdot z
\end{equation}
then the cross-section choice gives us a system of $r$ equations
\begin{equation}
w_j\left(g,z\right) = c_j ,~~ j=1,2,\ldots,r
\label{eq:exp}
\end{equation}
The system~\eqref{eq:exp} is solved for the group parameters $g=g\left(g_1,\ldots,g_r\right)$ and the right moving frame is ultimately given as 
\begin{equation}
g = \rho\left(z\right)
\end{equation}
This group element clearly maps each $z$ to the cross-section $K$ since the group parameters were chosen to do so. Note that $r$ of the coordinates of $g\cdot z$ are specified by the cross-section, and so there remain $m-r$ ``un-normalized'' coordinates, namely $w_{r+1}\left(g,z\right),\ldots,w_{m}\left(g,z\right)$. If we substitute the previously computed moving frame parameters $g = \rho\left(z\right)$ into these coordinates, we obtain the fundamental invariants, as stated in the following theorem.
\begin{thm}
Let $g=\rho\left(z\right)$ be the moving frame solution to the normalization equations~\eqref{eq:exp}. Then
$$ I_1\left(z\right)=w_{r+1}\left(\rho\left(z\right),z\right), ~\ldots ~, I_{m-r}\left(z\right) = w_m\left(\rho\left(z\right),z\right) $$
are a complete system of functionally independent invariants, called the fundamental invariants.
\label{th:fs}
\end{thm}
In this way, we have computed all of the coordinates of the map $z \mapsto \rho\left(z\right)\cdot z$ as
\begin{equation}
\rho\left(z\right)\cdot z = \left(c_1,\ldots,c_r,I_{1}\left(z\right),\ldots,I_{m-r}\left(z\right)\right)
\end{equation}
As the name implies, the fundamental invariants are very useful, as evidenced by the following theorem.
\begin{thm}
Any invariant $I\left(z\right)$ can be uniquely expressed as a function
$$ I\left(z\right) = H\left(I_1\left(z\right),\ldots,I_{m-r}\left(z\right)\right) $$
of the fundamental invariants.
\label{th:fi}
\end{thm}

Further, given any (scalar) function on our manifold, we can ``invariantize'' the function by composing it with the moving frame map, as described in the following theorem.
\begin{thm}
Given function $F:M\rightarrow \mR$, the invariantization of $F$ with respect to right moving frame $g=\rho\left(z\right)$ is the invariant function $I=\iota\left(F\right)$ defined as
\begin{equation}
I\left(z\right) = F\left(\rho\left(z\right)\cdot z\right)
\end{equation}
\end{thm}
In fact, the expression of invariants in terms of the fundamental invariants as in Theorem~\ref{th:fi} can be accomplished from the following Replacement Theorem.
\begin{thm}
If $I\left(z_1,\ldots,z_m\right)$ is any invariant, then
$$ \iota\left[ I\left( z_1,\ldots,z_m\right) \right]  = I \left(c_1,\ldots,c_r,I_1\left(z\right),\ldots,I_{m-r}\left(z\right)    \right) $$
\label{th:rt}
\end{thm}
That is, any invariant is easily rewritten in terms of the fundamental invariants, which serves to prove Theorem 4.

A major difficulty arises when we attempt to apply this construction to some common group actions: many group actions are not free, often resulting from the dimension of the manifold being less than the dimension of the group. In these cases, the normalization equations~\eqref{eq:exp} cannot be fully solved for the group parameters. Examples of non-free group actions include the Euclidean and affine groups on $\mR^2$. Theorem~\ref{th:ex} does not directly apply for non-free group actions. However, we may increase the dimension of the manifold by prolonging the group action to the jet spaces, that is, we consider instead the prolonged group action on the derivatives
\begin{equation}
G^{\left(n\right)} : J^{n}\left(M,p\right) \rightarrow J^n\left(M,p\right)
\end{equation}
Generally, if the action is prolonged to a high enough order jet space, the action will become (locally) free, \cite{olv1995}, and we can proceed with our construction; we will have enough equations to solve system~\eqref{eq:exp}. This prolongation can be accomplished by implicit differentiation, as demonstrated in the following section.

\section{Equi-Affine Invariants for 2D Images}
\label{sec:2d}
In this section we follow the second author's note \cite{olv2015a} to apply the above procedure to construct the 2D equi-affine differential invariants in the context of 2D image transformations. In particular, we are considering a grayscale image as a function $u:\Omega\subset \mR^2\rightarrow \mR$, and we seek the differential invariants of this function under equi-affine transformations of the domain coordinates, $\left(x,y\right)\in \Omega$. This expression in terms of image derivatives is useful since these can be computed (approximated) directly from image data.

The equi-affine group $\hbox{SA}(2)=\hbox{SL}(2)\ltimes \mR^2$ acts on plane coordinates $\left(x,y\right)$ as
\begin{equation}
\left( \begin{array}{c} z \\ w \end{array} \right) = \left( \begin{array}{cc} \alpha & \beta \\ \gamma & \delta \end{array} \right)\left( \begin{array}{c} x \\ y \end{array} \right) + \left( \begin{array}{c} a \\ b \end{array}\right),
\label{eq:eat}
\end{equation}
subject to the constraint 
$$\alpha\delta-\beta\gamma = 1.$$
 Each transformation in $\hbox{SA}\left(2\right)$ contains five independent parameters, thus $\hbox{SA}(2)$ is a five-dimensional Lie group. Note that these transformations are always invertible, with inverse
\begin{equation}
\left( \begin{array}{c} x \\ y \end{array} \right) = \left( \begin{array}{cc} \delta & -\beta \\ -\gamma & \alpha \end{array} \right)\left( \begin{array}{c} z-a \\ w-b \end{array} \right).
\label{eq:eati}
\end{equation}
We let $M = \mR^3$ with coordinates $x,y,u$, where the image function $u$ depends on $x,y$ and is not affected by the equiaffine group transformations.
Since $\dim G = 5 > \dim M = 3$, this group action is not free. 
Hence, Theorem~\ref{th:ex} does not apply, and we must prolong the group action to a higher-order jet space such that the action becomes free. For this example, we need only to prolong to the second jet space, which has dimension $8$. (Although the first order jet space has dimension $5$, the fact that $u$ is invariant means that the action is not free at order $1$.) We expect $8 - 5 = 3$ independent differential invariants, one of which is $u$ itself.

The prolonged action of the transformation indicates how the coordinate mapping~\eqref{eq:eat} will affect the image function $u$ and its derivatives. Clearly, the function $u$ is unchanged by a change of coordinates. Applying the chain rule of multivariate calculus, we can use the transformation~\eqref{eq:eat} to construct the prolonged derivatives, that is, how the derivatives of the ``transformed'' function relate to the original derivatives and the transformation parameters. For the $x$-coordinate,
\begin{equation}
u_z = u_x \frac{\partial x}{\partial z}+u_y \frac{\partial y}{\partial z} = \delta u_x - \gamma u_y
\label{eq:pr1}
\end{equation}
Similarly,
\begin{equation}
u_w = -\beta u_x + \alpha u_y
\label{eq:pr2}
\end{equation}
More generally, this differentiation can be represented by the implicit differentiation operators
$$ D_z = \delta D_x - \gamma D_y, ~~ D_w = -\beta D_x + \alpha D_y $$
where the $D$ represent total differentiation operators with respect to their subscripts.
Continuing in this fashion, we can construct the higher-order prolonged derivatives by repeated application of these operators. The second-order derivatives are 
\begin{equation}
\begin{array}{c} 
u_{zz} =  \delta^2 u_{xx}-2\gamma\delta u_{xy}+\gamma^2 u_{yy} \\
u_{ww} = \beta^2 u_{xx}-2\alpha\beta u_{xy} + \alpha^2 u_{yy} \\
u_{zw} = -\beta\delta u_{xx}+\left(\alpha\delta+\beta\gamma\right)u_{xy}-\alpha\gamma u_{yy}
\end{array}
\label{eq:pr3}
\end{equation}

The next step in the procedure is to normalize by choosing a cross-section to the group orbits. This amounts to equating several of expressions~\eqref{eq:eat}, \eqref{eq:pr1}, \eqref{eq:pr2}, and \eqref{eq:pr3} to constants. These constants must be chosen such that the resulting system is solvable for the group parameters. Since $\hbox{SA}(2)$ is five-dimensional, we need to choose five constants. The cross-section we will use is 
\begin{equation}
K = \left\{ ~z=w=u_z=u_{zw}=0, ~~ u_w = 1 ~\right\}
\label{eq:cs}
\end{equation}
as in~\cite{olv2015a}. This (nonlinear) system is readily solved for the group parameters. Notice that since we normalized $u_z = 0$ and $u_w=1$, we obtain
\begin{equation}
\delta = \gamma \frac{u_y}{u_x} \qquad {\rm and} \qquad \alpha = \frac{1+\beta u_x}{u_y},
\label{d:d}
\end{equation}
respectively. Combining these with the unimodality constraint, $\alpha\delta-\beta\gamma=1$ and then using expression~\eqref{d:d}, we find
\begin{equation}
\gamma = u_x \qquad {\rm and} \qquad \delta = u_y
\end{equation}
Substituting these relationships into the second-order equation $u_{zw}=1$, we can solve for $\beta$ as
\begin{equation}
\beta = \frac{u_{y}u_{xy}-u_{x}u_{yy}}{u_x^2u_{yy}-2u_xu_yu_{xy}+u_y^2u_{xx}}
\end{equation}
With this in hand, we compute $\alpha$ as 
\begin{equation}
\alpha = \frac{u_yu_{xx}-u_xu_{xy}}{u_x^2u_{yy}-2u_xu_yu_{xy}+u_y^2u_{xx}}
\end{equation}
The translation parameters $a$ and $b$ can be computed easily from the choices $z=w=0$ as
\begin{equation}
a = -\alpha x - \beta y  \qquad {\rm and} \qquad 
b = -\gamma x - \delta y
\end{equation}
where the values found for the other parameters may be substituted in. Taken together, these group parameters define the right moving frame corresponding to our cross-section $K$.

Having computed this, we are now in a position to compute invariant objects. In particular, given any function $F$ of $u$ and its derivatives, we can invariantize $F$ by transforming it under~\eqref{eq:eat} and then replacing all instances of the group parameters by their moving frame expressions. Some natural functions to transform are those used in the cross-section normalization. For example, let $F=u_x$ and notice
$$ u_x \mapsto u_z =  \delta u_x -\gamma u_y $$
Plugging in the above expressions, 
\begin{equation}
\iota \left(u_x\right) = u_y u_x - u_x u_y = 0
\end{equation}
This is obvious, of course, since the expressions for the group parameters were chosen according to the cross-section in which we insisted $u_z=0$. The invariantized forms of the functions used in the cross-section normalization are constant, and are known as the phantom invariants.

More interestingly, Theorem~\ref{th:fs} guarantees the existence of a set of fundamental second-order invariants. Consider the second derivatives $u_{zz}$ and $u_{ww}$ which were not used in the normalization. Invariantizing these, we obtain the well-known second order equiaffine differential invariants:
\begin{equation}
\iota\left(u_{zz}\right) = u_y^2u_{xx}-2u_xu_yu_{xy}+u_x^2u_{yy} = J
\label{eq:j}
\end{equation}
and
\begin{equation}
\iota\left(u_{ww}\right) = \frac{u_{xx}u_{yy}-u_{xy}^2}{J} = \frac{H}{J}
\label{eq:hj}
\end{equation}
According to Theorem~\ref{th:fi}, $H$ and $J$ form a complete system of second-order differential invariants. It can be checked directly via chain rule that these are truly equi-affine invariant. 

\begin{figure}[t]
\centering
\subfloat[]{\includegraphics[scale=.1]{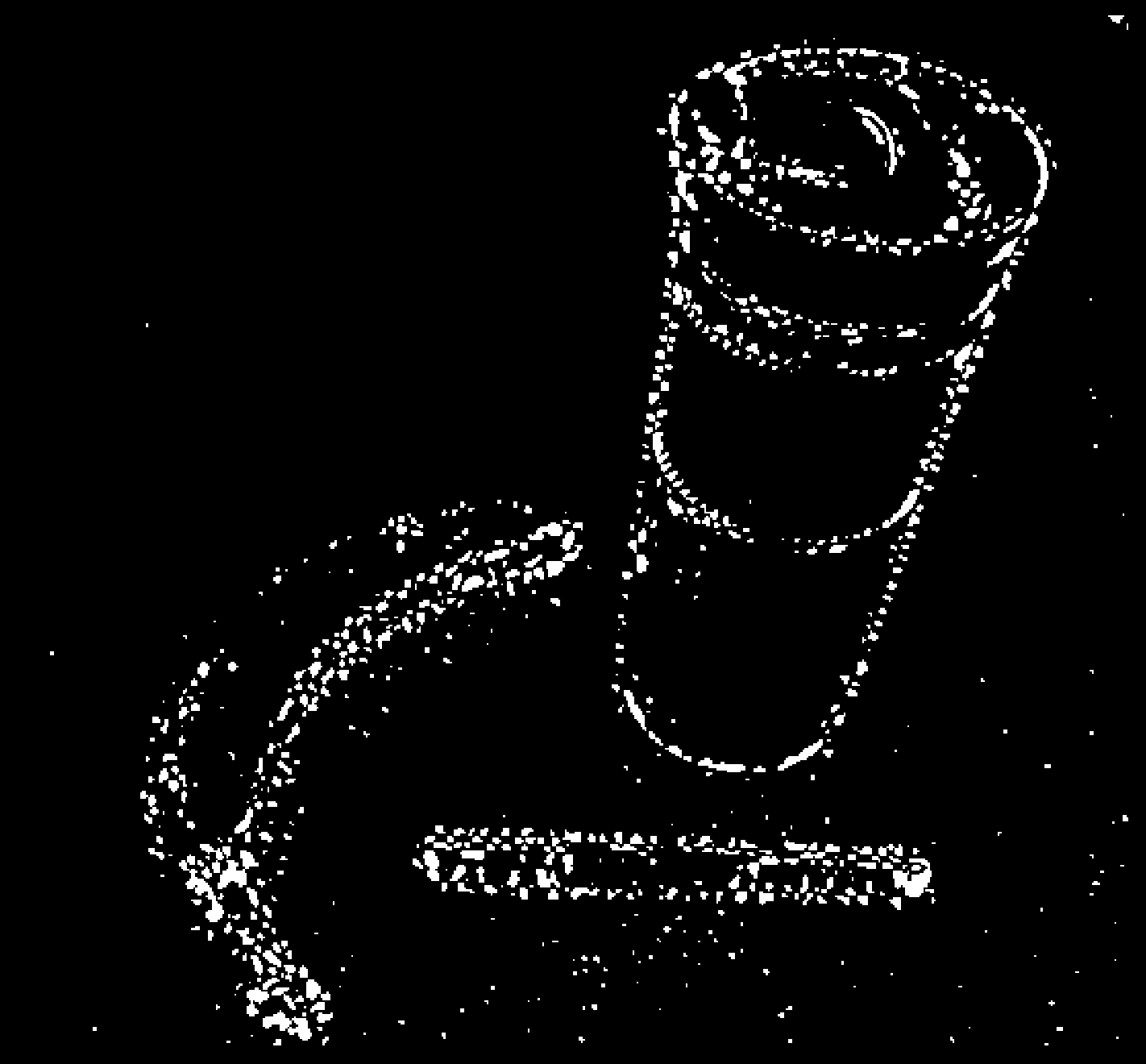}} ~~
\subfloat[]{\includegraphics[scale=.1]{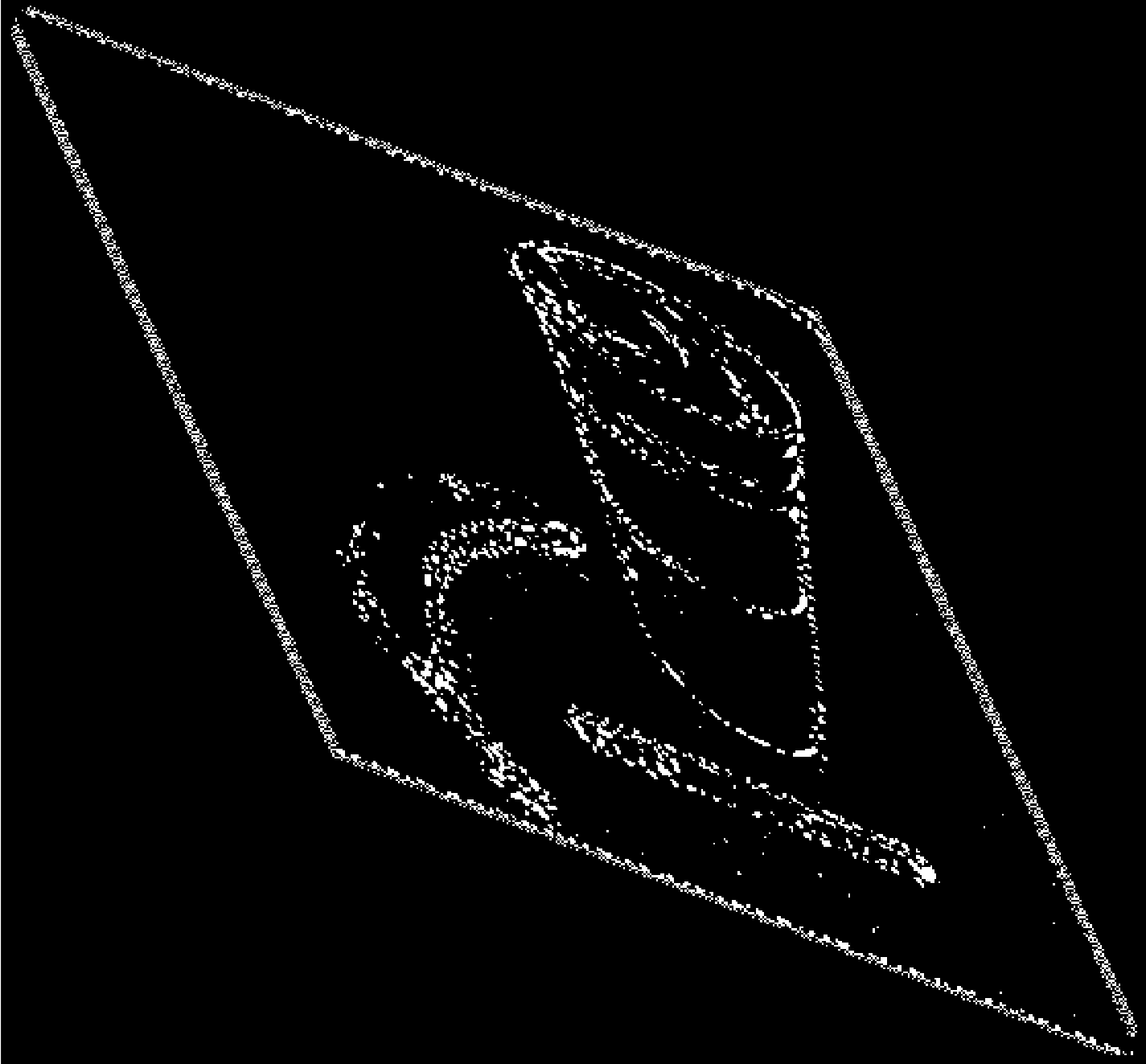}}\\
\subfloat[]{\includegraphics[scale=.1]{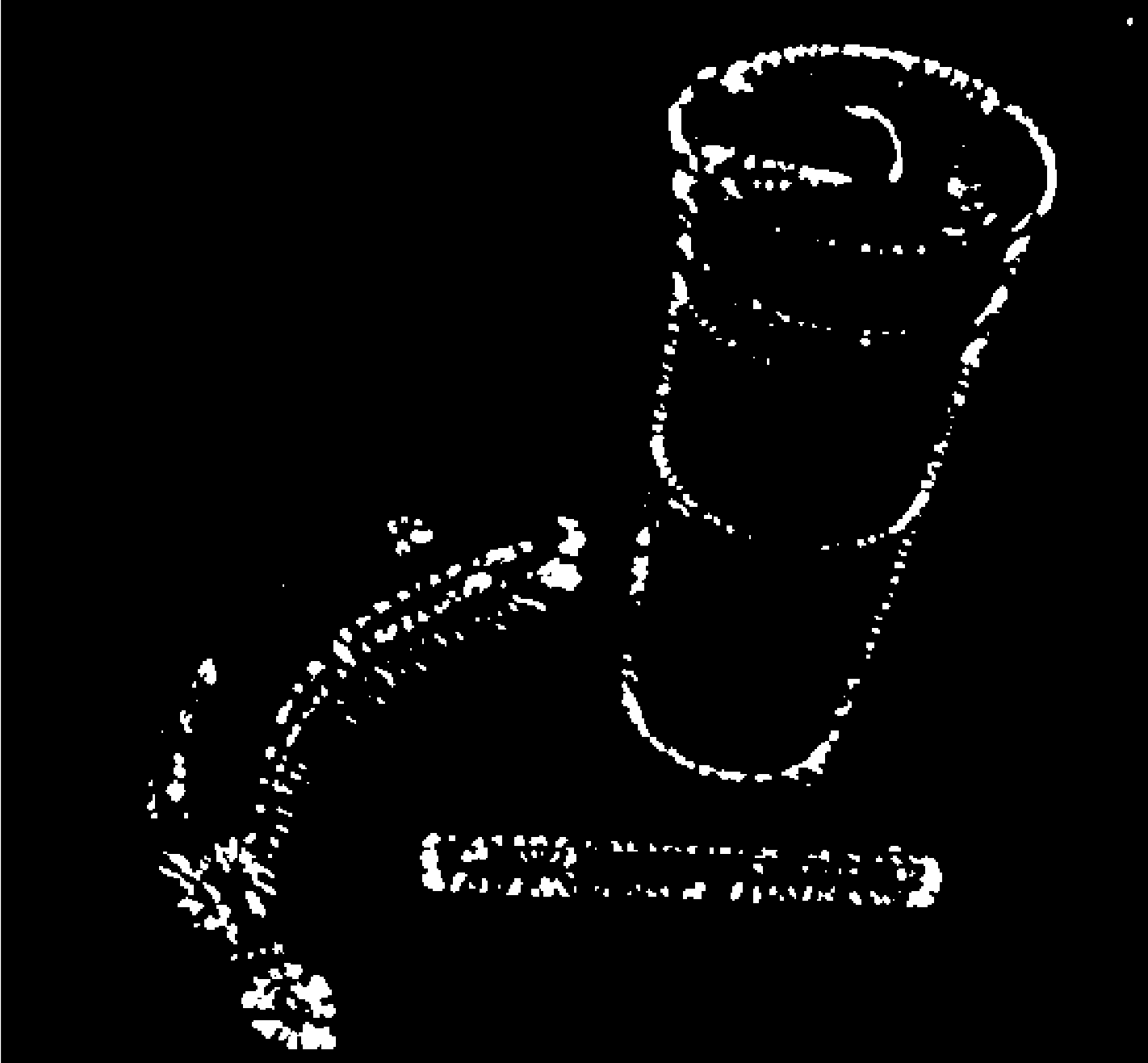}} ~~
\subfloat[]{\includegraphics[scale=.1]{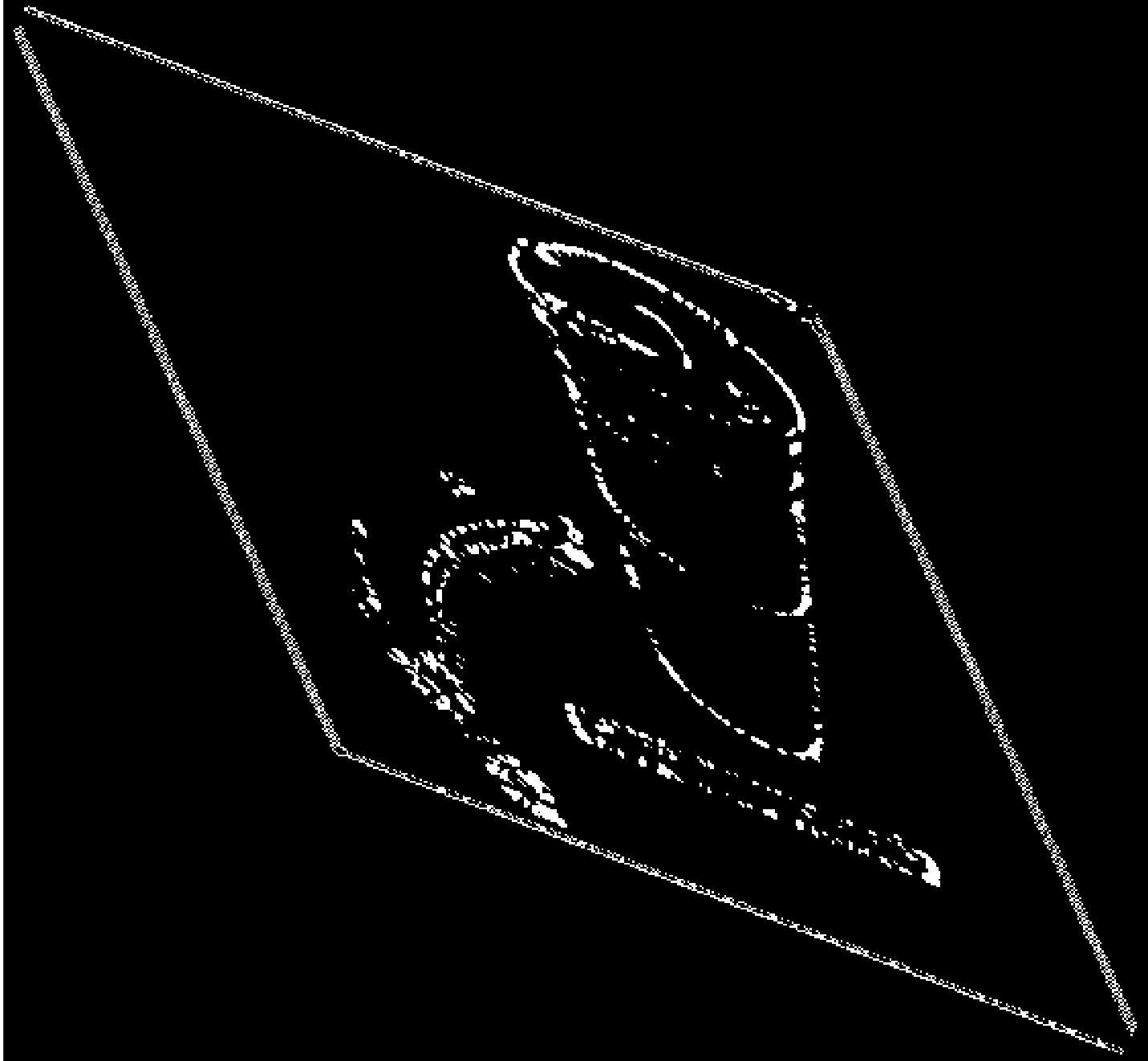}}
 \caption{Thresholded differential invariants $H$ (top) and $J$ (bottom) for the images in Figure~\ref{fig:ims}.}
 \label{fig:hex}
\end{figure}

Figure~\ref{fig:hex} demonstrates $H$ and $J$ computed on the image pair in Figure~\ref{fig:ims}. These functions are thresholded for clarity, but it can be observed that they are generally unchanged by the transformation. Differences can be attributed to the fact that our derivative computations via finite differences are computed on a Euclidean grid and are thus not affine-invariant themselves.

\emph{Remark\/}: Moving frames can be used to design group-invariant numerical approximations to differential invariants and invariant differential equations, as in~\cite{cal1996,wel2007}. A future project is to develop equi-affine finite difference approximations to the differential invariants, which would better serve to localize invariant feature points.

\subsection{The 3D Case}
The extension to 3D image volumes has recently been investigated. Fully three-dimensional image data has become more popular recently, for example, in the medical field. Here we describe the computation of differential invariants of the equi-affine group acting on three-dimensional images.

The equi-affine group $\hbox{SA}(3)=\hbox{SL}(3)\ltimes \mR^3$ acts on space coordinates $\left(x,y,z\right)$ as
\begin{equation}
\left( \begin{array}{c} u \\ v \\ w \end{array} \right) = \left( \begin{array}{ccc} \alpha & \beta & \gamma \\ \delta & \epsilon & \zeta \\  \eta & \theta & \lambda \end{array} \right)\left( \begin{array}{c} x \\ y \\ z \end{array} \right) + \left( \begin{array}{c} a \\ b \\ c \end{array}\right) = A \left( \begin{array}{c} x \\ y \\ z \end{array} \right) + \vec{a},
\label{eq:eat3}
\end{equation}
subject to the constraint $\hbox{det}\left(A\right) = 1$. Clearly, $SA(3)$ is an 11-dimensional Lie group since there are 11 independent parameters. As in the two-dimensional case, this group action is certainly not free, and we must prolong the group action to a higher-order jet space such that the action becomes free. Unfortunately, when we prolong to the second jet space, $J^2$, the group action is still not free, even though the dimensions satisfy
$$ \hbox{dim }J^2 = 13 > 11 = \hbox{dim }G $$
Through our construction we will produce 3 functionally independent second-order differential invariants, and this proves that the (generic) orbits are 10-dimensional, which precludes freeness of the second order prolonged action.

We thus need to prolong the group action to the third-order jet space in order to find a usable cross-section. The cross-section we will employ is
\begin{equation}
K = \left\{~ x=y=z=0,~~u_x=u_y=u_{xy}=u_{yz}=u_{zz}=u_{zzz}=0,~~u_z=u_{xz}=1 ~\right\}
\label{eq:cs3}
\end{equation}
Notice that the variables $u_{xx}$ and $u_{yy}$ are not involved in the normalization, and so the invariantization of these quantities will give our desired fundamental second-order invariants.

The next step in the procedure would be to solve the system~\eqref{eq:cs3}. In this case, this is a nonlinear system of 11 equations, and computing the closed-form solution is not as straightforward as it was in the 2D case. Instead, we will extend the 2D differential invariants $H$ and $J$ to the 3D case and then apply the Replacement Theorem~\ref{th:rt}. Recall that this theorem allows us to express any invariant in terms of the moving frame invariants. From the cross-section~\eqref{eq:cs3}, we see
\begin{equation}
\iota \left( \nabla u\right) = \left( \begin{array}{c} 0 \\ 0 \\ 1 \end{array} \right)
\label{eq:invgrad}
\end{equation}
and
\begin{equation}
\iota \left( \nabla^2 u\right) = \left( \begin{array}{ccc} \iota\left(u_{xx}\right) & 0 & 1 \\ 0 & \iota\left(u_{yy}\right) & 0 \\ 1 & 0 & 0 \end{array} \right) = \left( \begin{array}{ccc} L & 0 & 1 \\ 0 & K & 0 \\ 1 & 0 & 0 \end{array} \right)
\label{eq:invhes}
\end{equation}
where $L=\iota\left(u_{xx}\right)$ and $K=\iota\left(u_{yy}\right)$, not used in the normalization, are the fundamental invariants we wish to determine.

Notice that the invariants $H$ and $J$ found in the 2D case can be written in more general forms:
\begin{equation}
H = u_{xx}u_{yy}-u_{xy}^2 = \hbox{det}\left( \nabla^2 u \right)
\end{equation}
and
\begin{equation}
J/H = \frac{1}{\hbox{det}\left(\nabla^2u\right)} \left(u_x^2u_{yy}-2u_xu_yu_{xy}+u_y^2u_{xx}\right)=\nabla u^T \left(\nabla^2 u\right)^{-1}\nabla u
\end{equation}
If we consider now extending these to the 3D case, where $u=u\left(x,y,z\right)$, the invariantization results~\eqref{eq:invgrad} and~\eqref{eq:invhes} give
\begin{equation}
H = \iota\left( \hbox{det}\left(\nabla^2u\right)\right) = -K
 \qquad {\rm and} \qquad 
 J/H = \iota\left( \nabla u^T \left(\nabla^2u\right)^{-1}\nabla u\right) = -L
\end{equation}
That is, we can compute the fundamental invariants $L$ and $K$ for the 3D case directly as a generalization of the 2D fundamental invariants $J/H$ and $H$, respectively. Hence, the fundamental invariants are now expressed as
\begin{equation}
\iota\left(u_{yy}\right) = \hbox{det}\left(\nabla^2 u\right) = H
\end{equation}
and
\begin{equation}
\iota\left(u_{xx}\right) = \nabla u^T \left(\nabla^2 u\right)^{-1}\nabla u = J/H
\end{equation}
which can be computed directly from the image.

\begin{figure}[t]
\centering
\subfloat[]{\includegraphics[scale=.1]{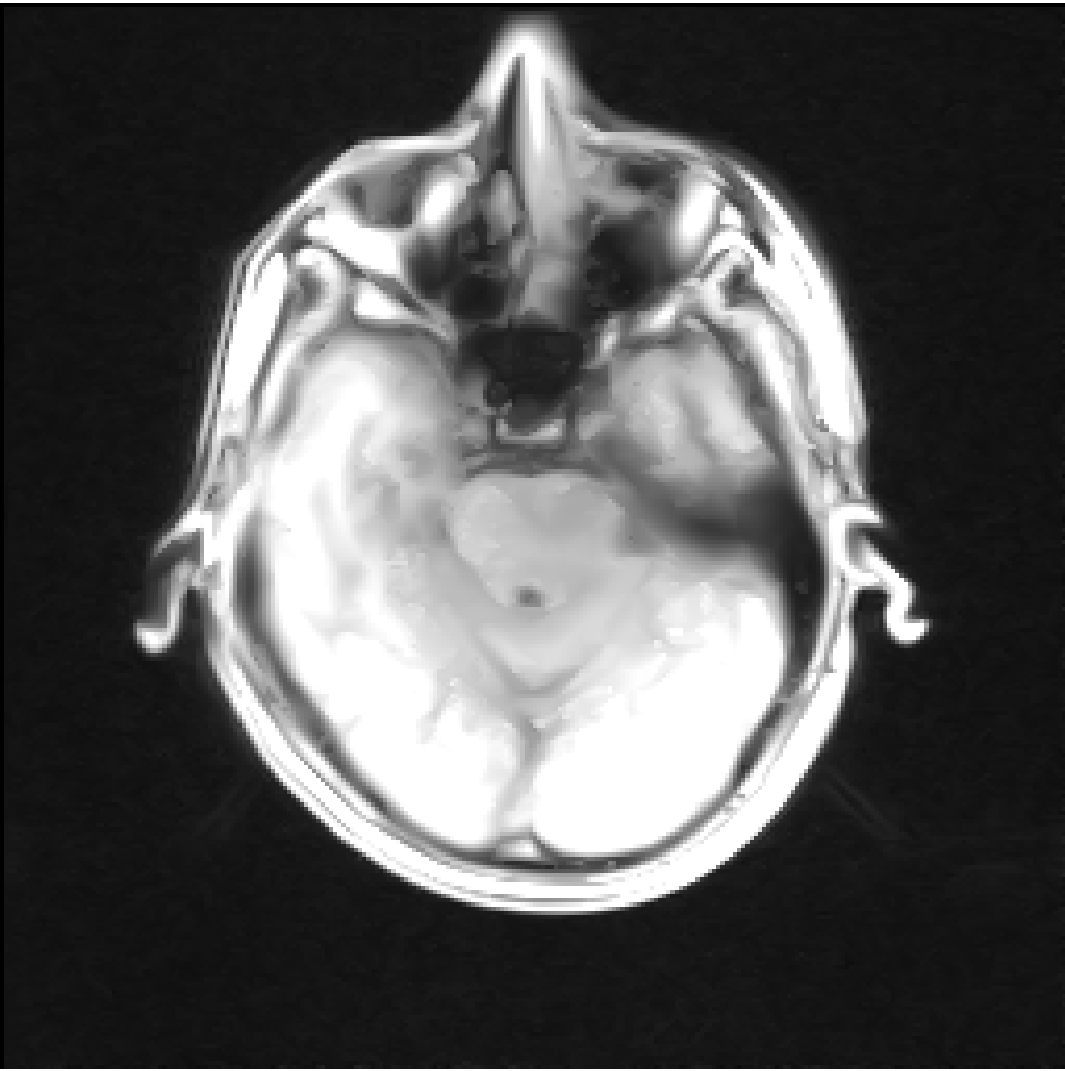}} ~~
\subfloat[]{\includegraphics[scale=.1]{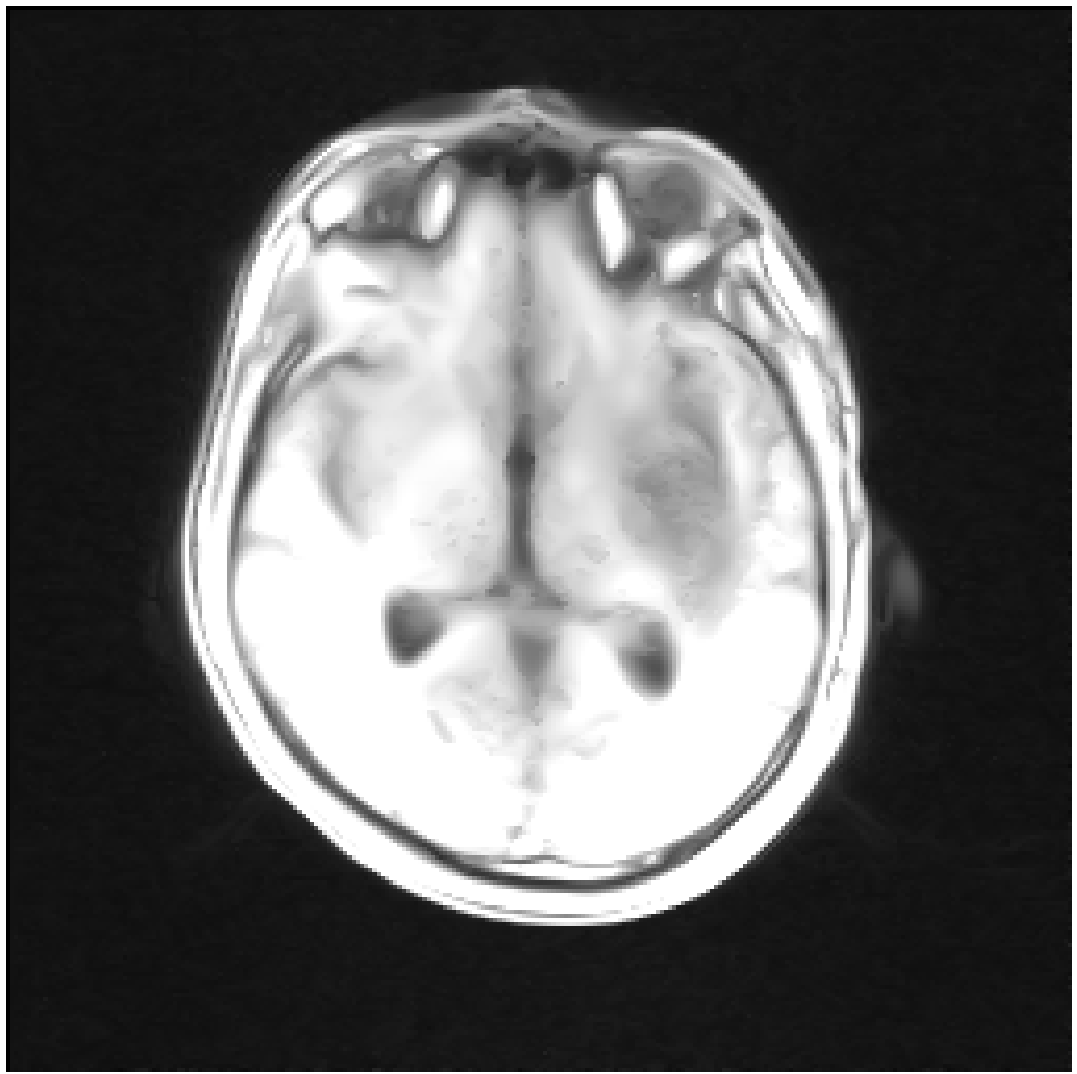}} ~~
\subfloat[]{\includegraphics[scale=.1]{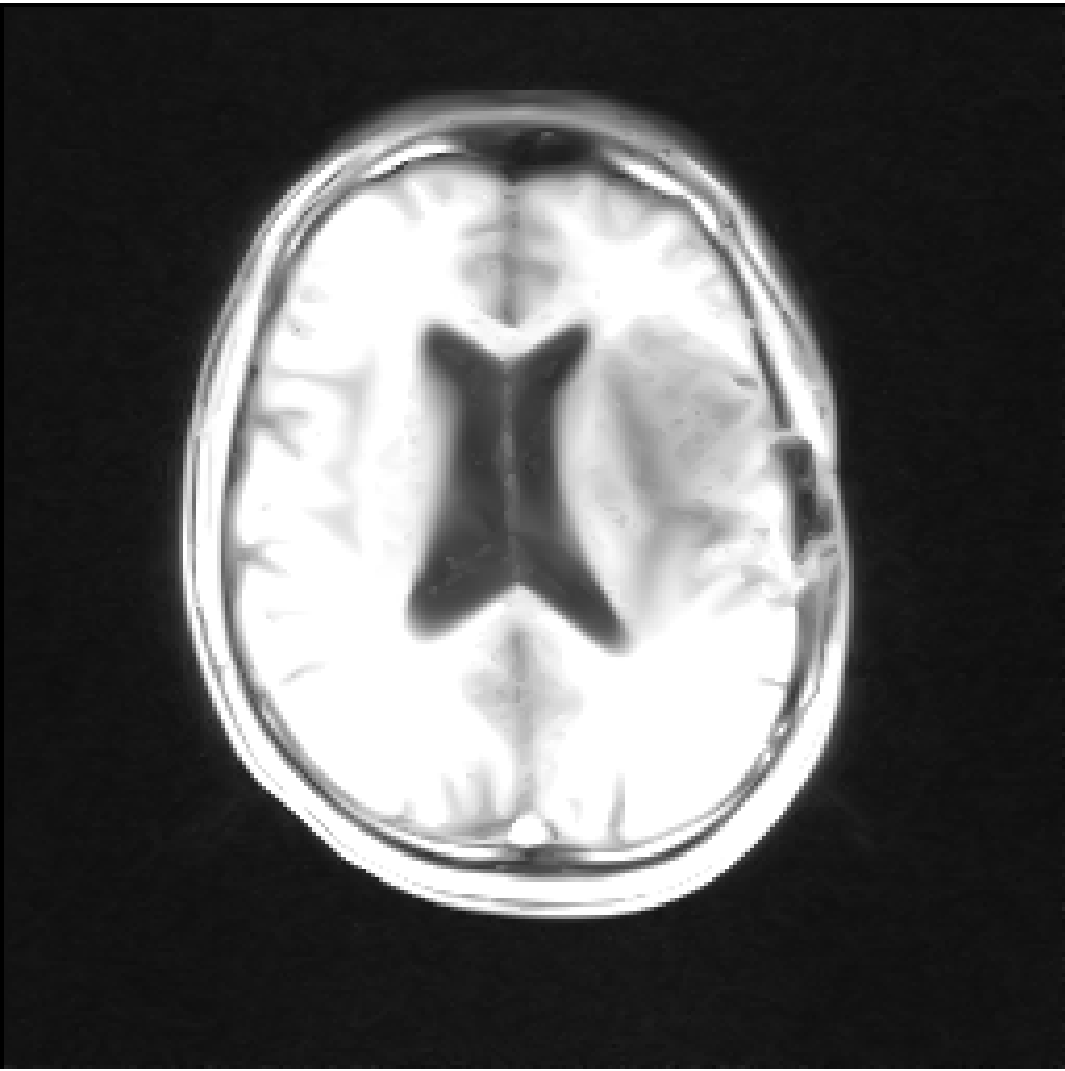}}\\
\subfloat[]{\includegraphics[scale=.1]{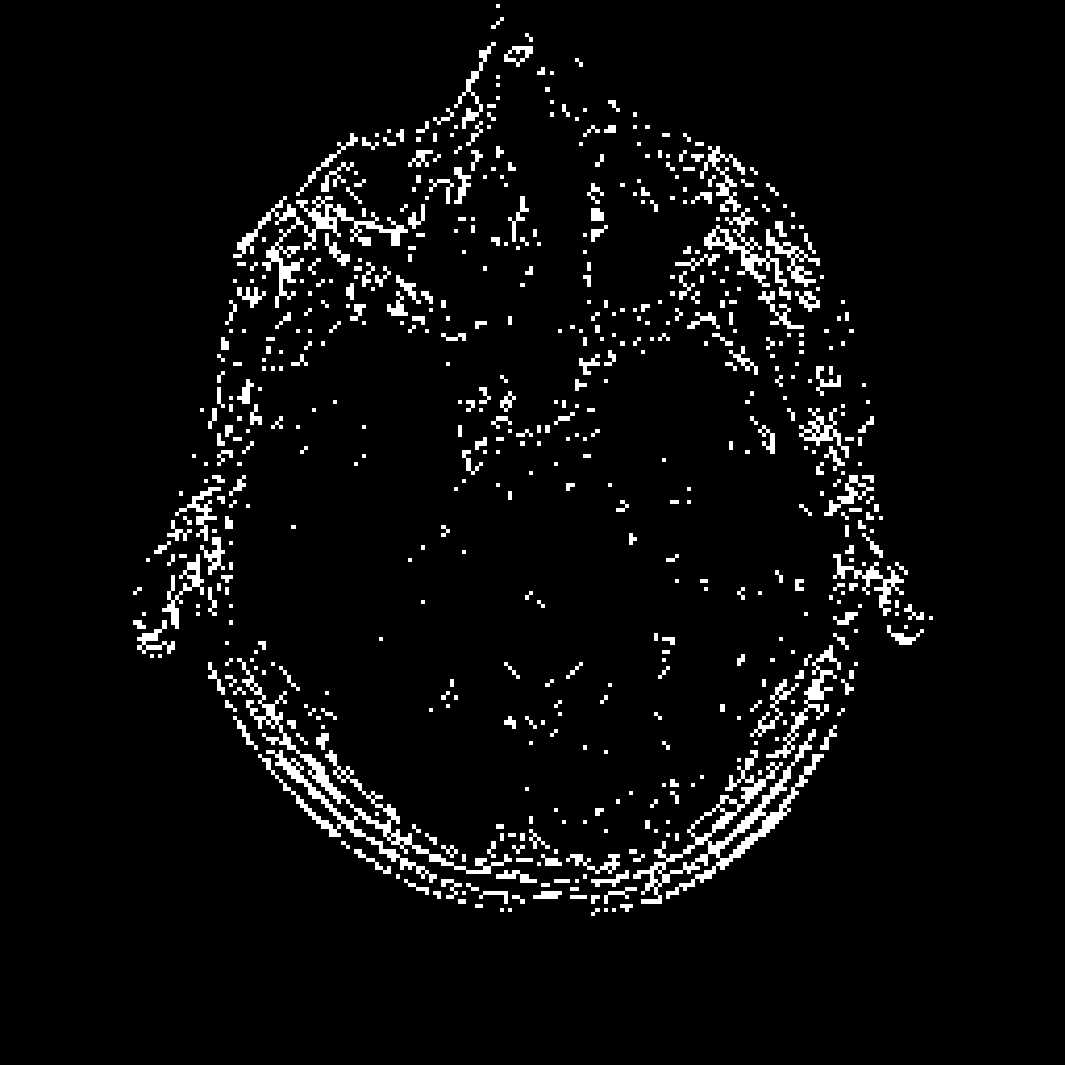}} ~~
\subfloat[]{\includegraphics[scale=.1]{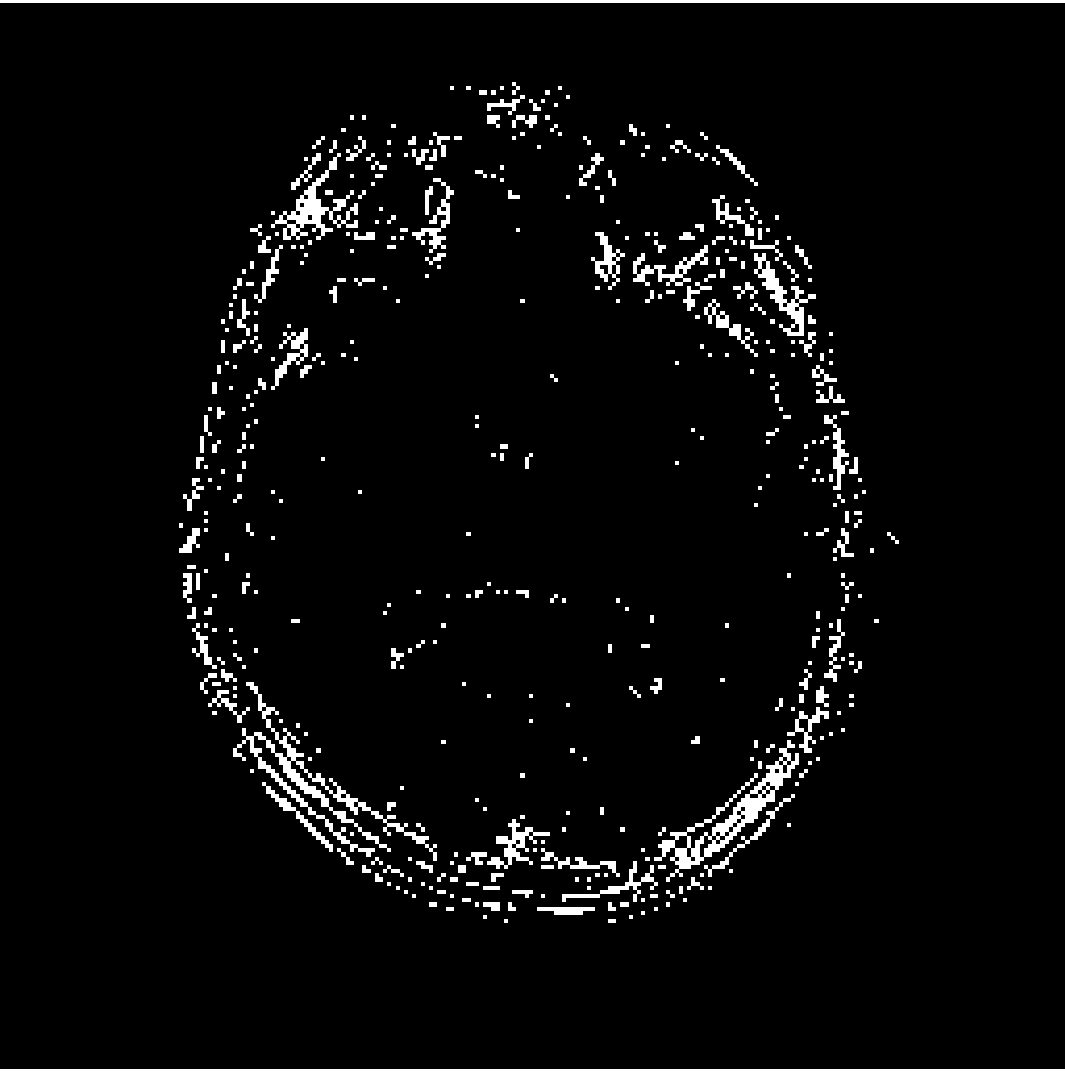}} ~~
\subfloat[]{\includegraphics[scale=.1]{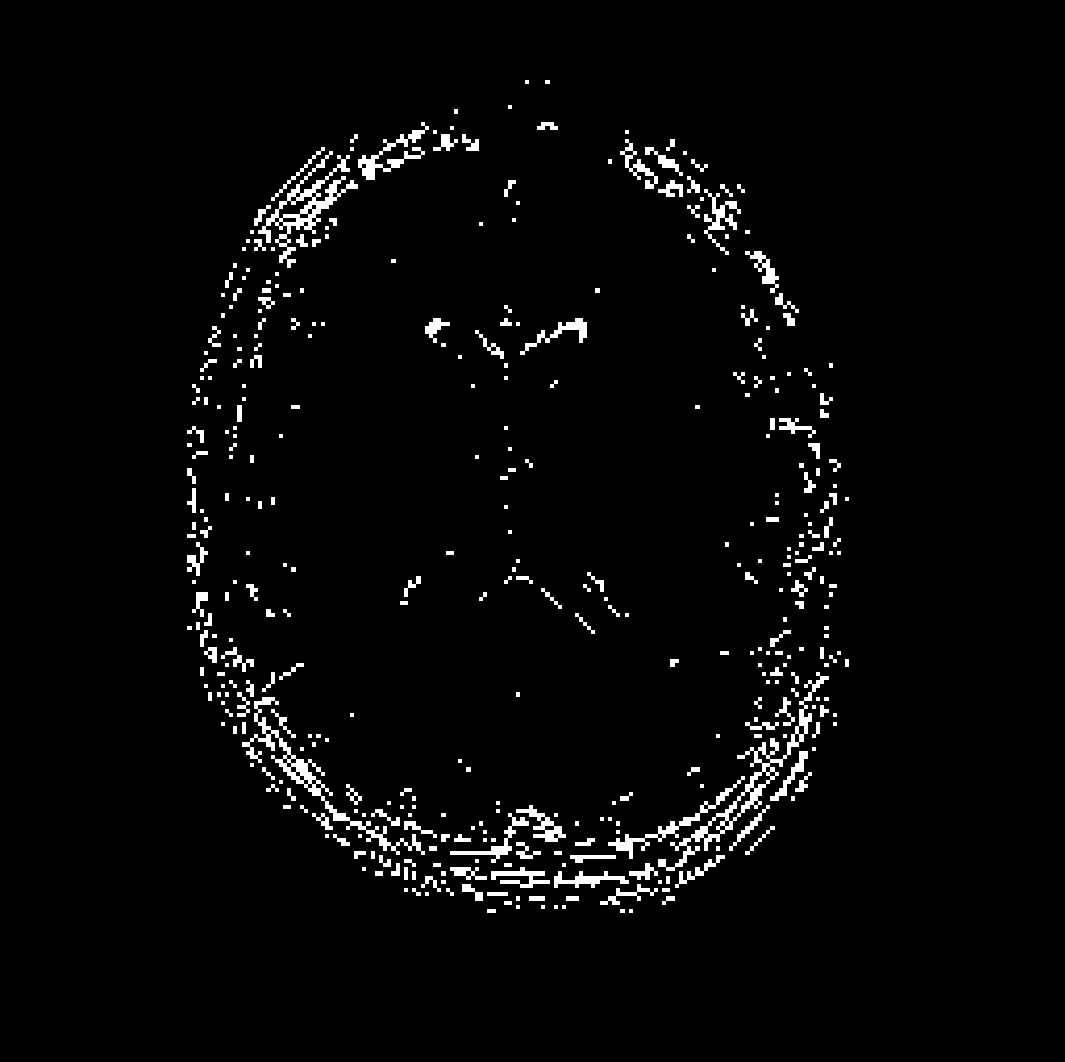}}\\
\subfloat[]{\includegraphics[scale=.1]{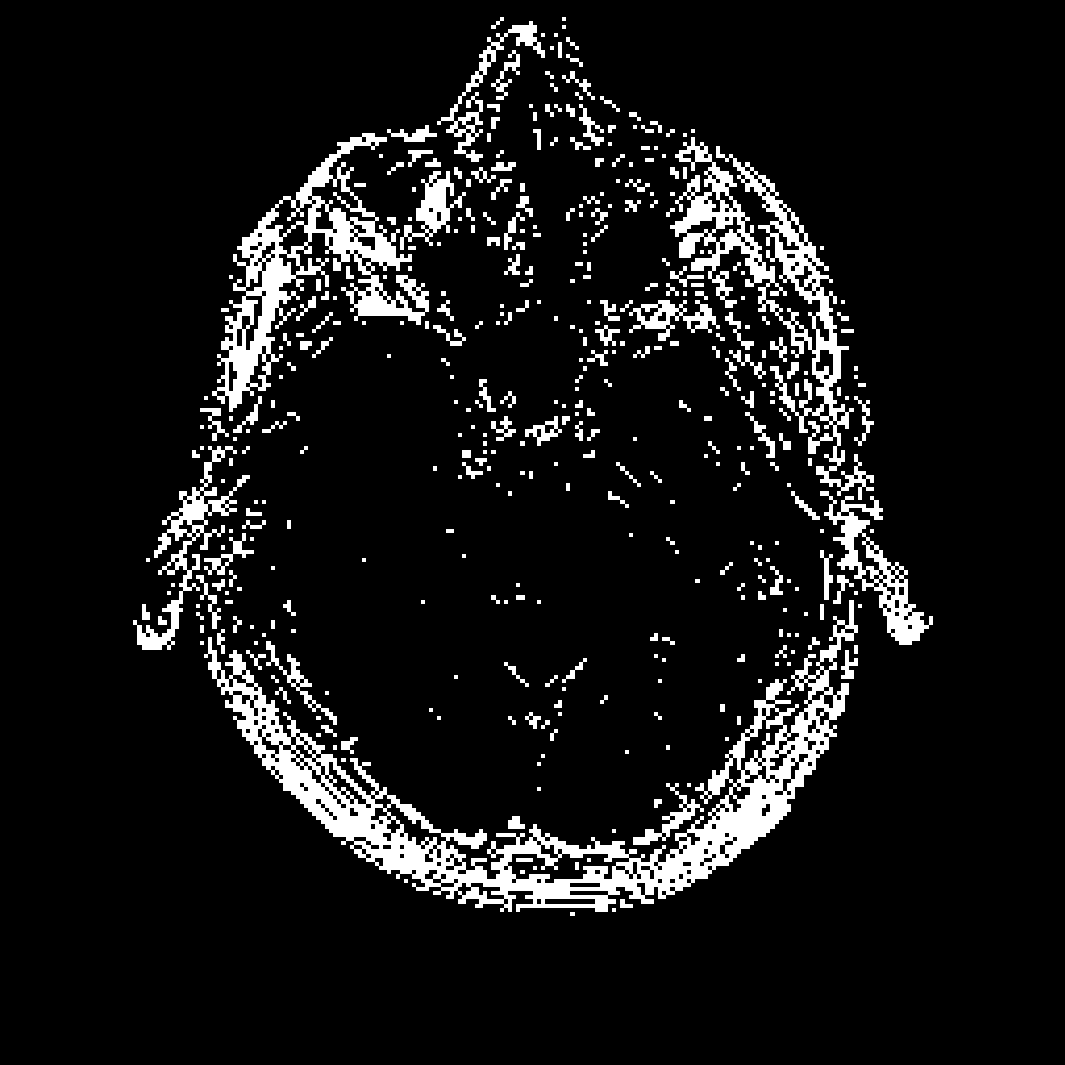}} ~~
\subfloat[]{\includegraphics[scale=.1]{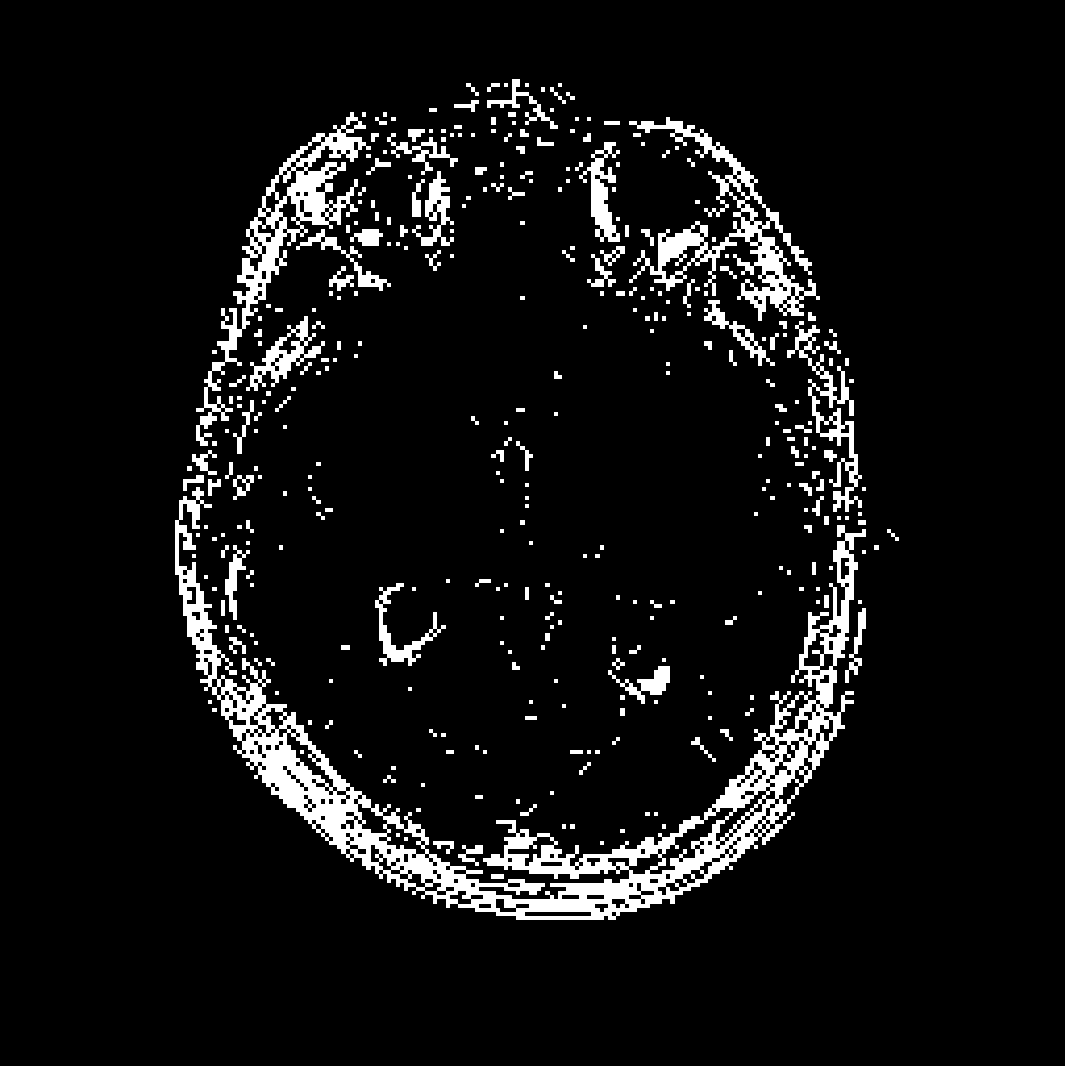}} ~~
\subfloat[]{\includegraphics[scale=.1]{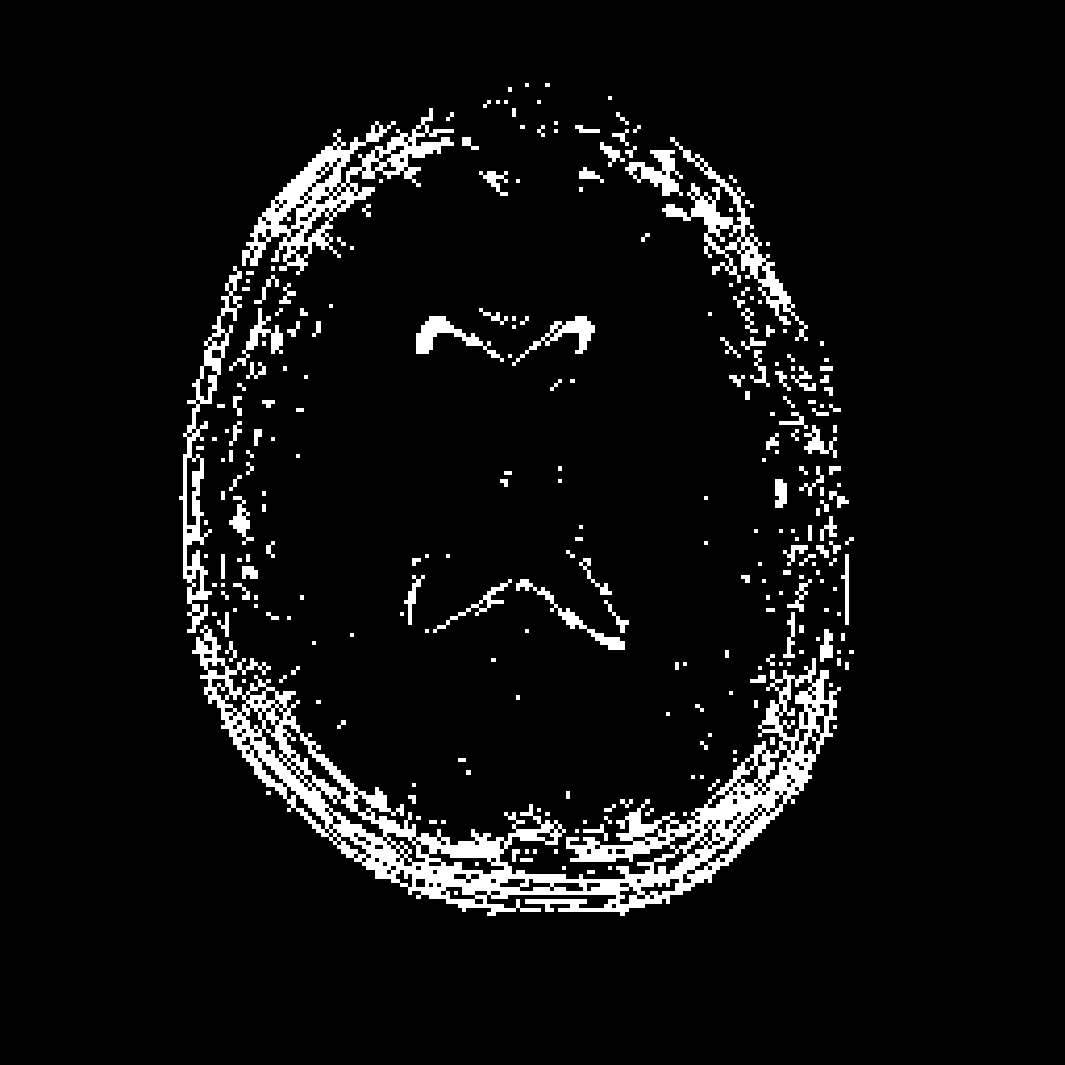}}
 \caption{Thresholded differential invariants $H$ (top) and $J$ (bottom) for a 3D image of a human brain: image slices (top row), $H$ (middle row), and $J$ (bottom row).}
 \label{fig:hex3}
\end{figure}

In Figure~\ref{fig:hex3}, we show these invariants for a typical 3D image. Several slices of an image volume are shown, along with the differential invariants computed on these slices. The exhibit similar characteristics as in the 2D case, being clustered about edges and corners, and are equi-affine invariant.

Notice that we do not need to compute any third-order differential invariants. However, these can be generated from the second-order differential invariants by invariant differentiation, but are omitted for space since the computations prove somewhat lengthy.

\section{Application: 2D Feature Detection with Scale-Space}
\label{sec:fd}
In this section, the differential invariants $H$ and $J$ are combined into an affine-invariant function which will serve as a feature detector. To make detection more robust to noise, we construct a scale-space of our images. Unlike traditional methods such as SIFT~\cite{low2004}, our scale-space is nonlinear, affine-invariant, and based on a geometric curve evolution equation.

\subsection{Affine-Invariant Gradient Magnitude}
\begin{figure}[t]
\centering
\subfloat[]{\includegraphics[scale=.1]{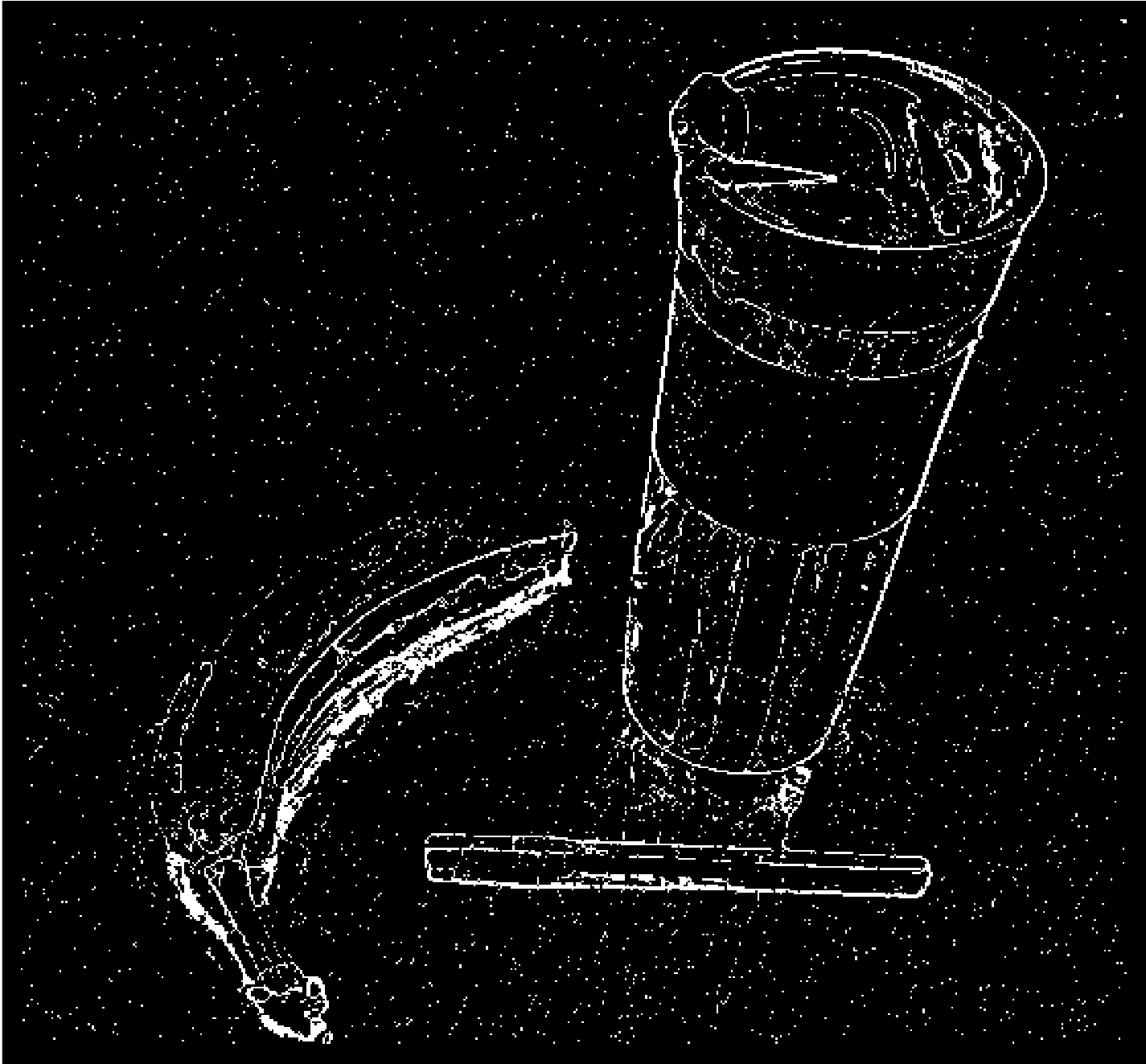}} ~~
\subfloat[]{\includegraphics[scale=.1]{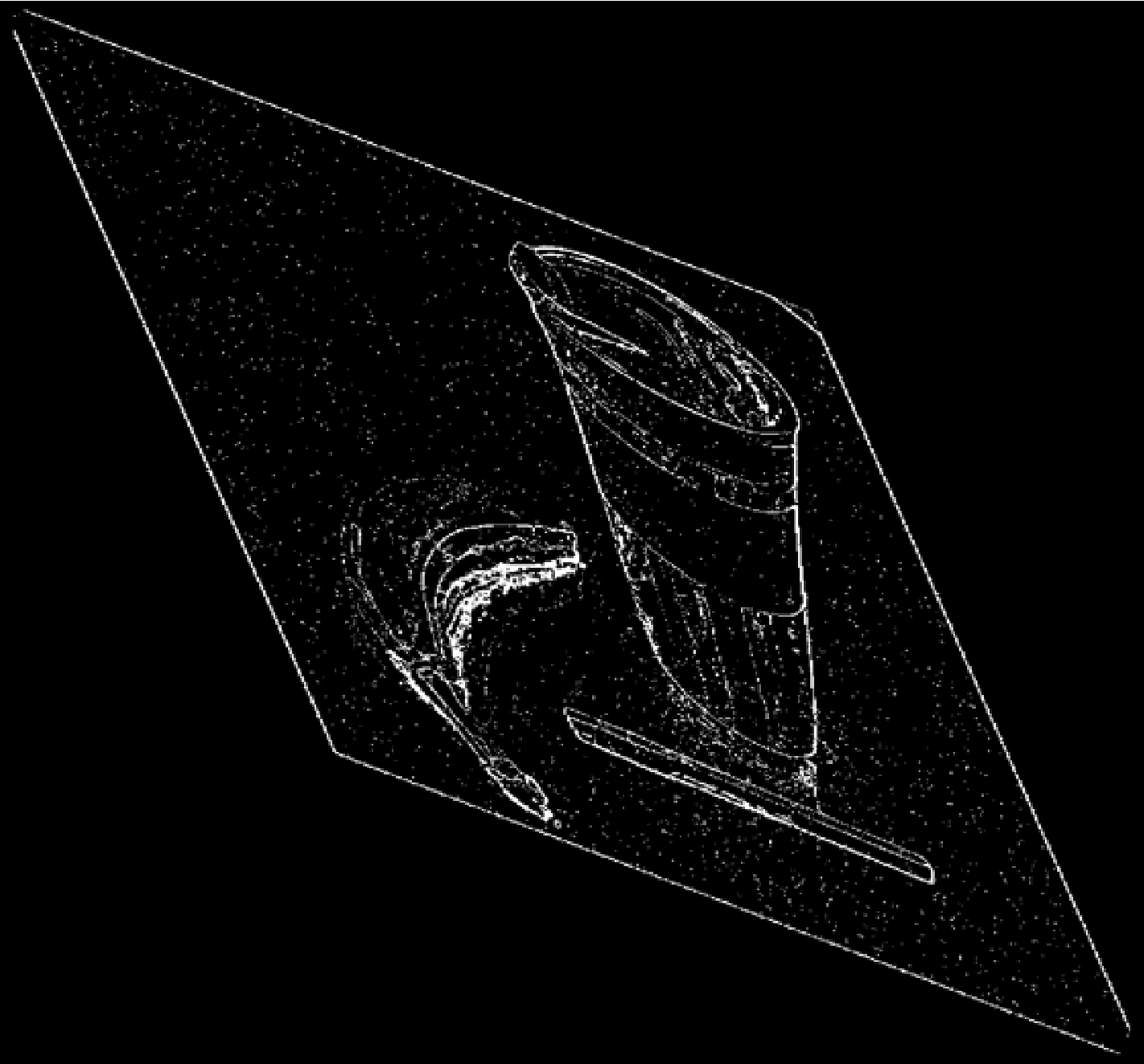}}
 \caption{Affine-invariant gradient magnitudes of the images in Figure~\ref{fig:ims}, thresholded for clarity.}
 \label{fig:aig}
\end{figure}

The affine differential invariants computed above were used in \cite{olv1999} to construct an affine-invariant analog to the traditional Euclidean gradient magnitude, namely
\begin{equation}
\nabla_{\hbox{aff}} u = \left| \frac{H}{J} \right|
\end{equation}
This function is slightly modified to avoid zero divisors:
\begin{equation}
\widehat{ \nabla}_{\hbox{aff}}   u = \sqrt{ \frac{H^2}{J^2+1} }
\label{eq:det}
\end{equation}
An example of this function applied to the image pair in Figure~\ref{fig:ims} is shown in Figure~\ref{fig:aig}. This function was previously applied as an edge detector for an affine-invariant active contour model. Indeed, the function can be seen to give large response along object boundaries.

Some example feature points identified as local maxima of the function~\eqref{eq:det} are shown in Figures~\ref{fig:ex1},~\ref{fig:ex2}, and~\ref{fig:ex3}. As expected, these points largely cluster around object boundaries. This poses a difficulty, since edge points are not distinct and tend to foil matching algorithms. A further modification may be made such that, for example, candidate feature points are only retained as true feature points if image variation is significant in multiple directions.

\subsection{Affine-Invariant Scale-Space}
\begin{figure}[h]
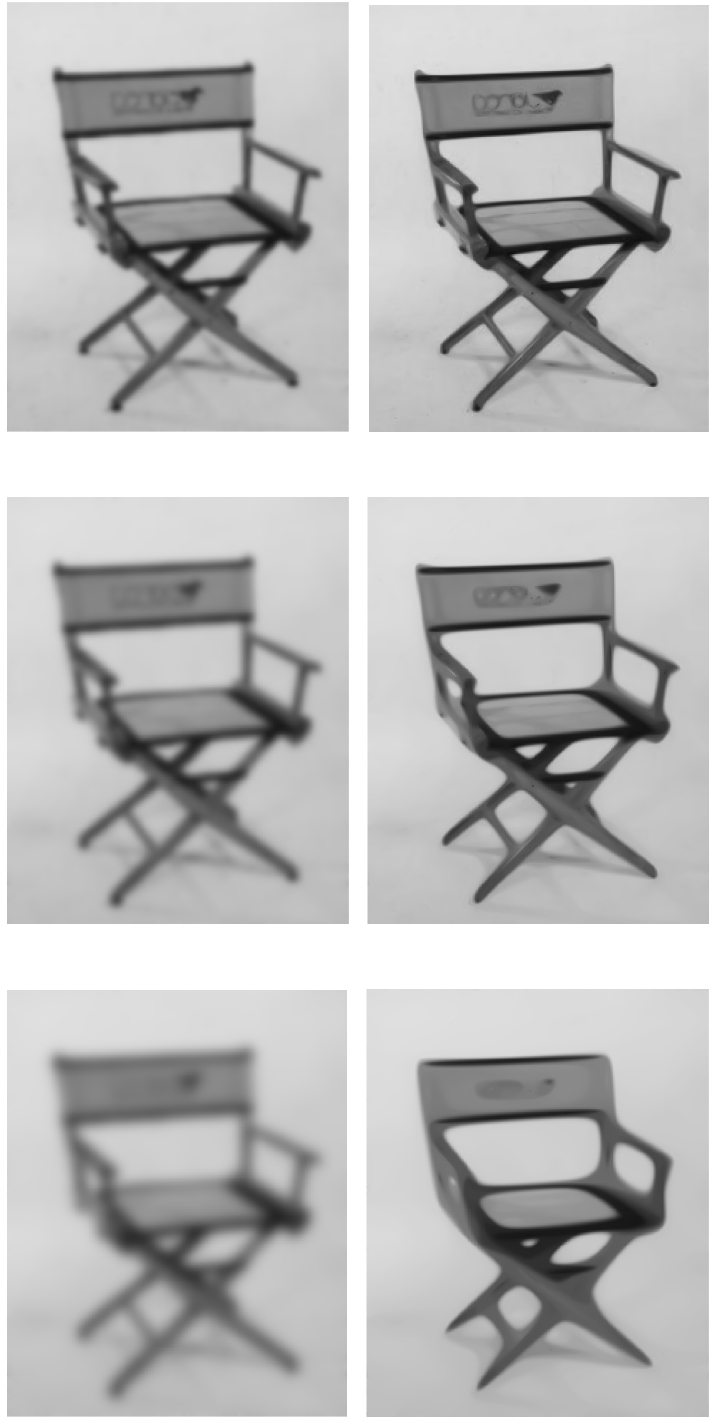

\captionsetup[subfigure]{labelformat=empty}
\centering
\subfloat[]{\includegraphics[scale=.14]{g1.png}} ~~
\subfloat[]{\includegraphics[scale=.14]{a1.png}} \\
\subfloat[]{\includegraphics[scale=.14]{g2.png}} ~~
\subfloat[]{\includegraphics[scale=.14]{a2.png}} \\
\subfloat[]{\includegraphics[scale=.14]{g3.png}} ~~
\subfloat[]{\includegraphics[scale=.14]{a3.png}}
 \caption{First column: samples of linear (Gaussian) scale-space of an image. Second column: samples of the affine-invariant scale-space of the same image. Scale parameter $t$ increases as we travel down the columns.}
 \label{fig:aiss}
\end{figure}

An important pre-processing step for feature detection is the construction of the Gaussian (or linear) scale-space of the image intensity signal by convolution of the image with Gaussian kernels of increasing standard deviation or, equivalently, evolution of the image via the linear diffusion equation
\begin{equation}
u_t = \Delta u 
\label{eq:linear}
\end{equation}
with initial condition $u\left(x,0\right)=u_0\left(x\right)$, where $u_0:\Omega\subset \mR^2 \rightarrow \mR$ is the original image, $\Delta$ is the usual Laplacian operator, and $t$ is an artificial time parameter. The solution to this equation, $u\left(x,t\right)$, for $x\in \mR^2$ and $t\in \left[0,\infty \right)$ is the Gaussian scale-space of the image \cite{wit1984}. This is a multi-scale representation of the image, a one-parameter family of images $u\left(\cdot,t\right)$ with the characteristic that as $t$ is increased, noise and high-frequency features are removed and we obtain successively smoothed versions of the original image data \cite{rom2013}. Feature points can then be extracted from this scale-space image representation by finding local detector maxima in both space and scale~\cite{low2004}. By searching across all scale-space, feature points can be identified at a variety of scales. A notable drawback of this linear approach is that important image features such as edges and corners are blurred by the smoothing, and so feature localization is lost.

Nonlinear scale-spaces have also been investigated for feature detection. The popular KAZE algorithm \cite{alc2012} relies on the Perona-Malik anisotropic diffusion framework in which an image scale-space is constructed as the solution to the nonlinear diffusion equation
\begin{equation}
\Phi_t = \hbox{div}\left( g\cdot \nabla \Phi \right)
\label{eq:perona}
\end{equation}
where $g$ is a conductance function which serves to slow diffusion near edges in an image \cite{per1990}. This image representation preserves edges better than its linear counterpart \eqref{eq:linear}, which is obtained from \eqref{eq:perona} when $g \equiv 1$.

For a fully affine-invariant multi-scale feature detection pipeline, we require both an affine-invariant scale-space image representation and an affine-invariant feature point detector. A fully affine-invariant scale-space for plane curves was developed in~\cite{sap1993}. In this work, a closed plane curve $C_0:\left[ 0,1\right]\rightarrow \mR^2$ evolves according to the partial differential equation (PDE)
\begin{equation}
\frac{\partial C}{\partial t} = \kappa^{1/3}N
\label{eq:ce}
\end{equation}
where $t$ is an artificial time parameter, $\kappa$ is the Euclidean curvature of $C$, and $N$ is the inward unit normal vector to $C$. The solution to this equation is a one-parameter family of plane curves, $C\left(\cdot,t\right)$, which evolve in an affine-invariant manner to smooth out the curve. By evolving the PDE for larger amount of time, we obtain successively smoothed versions of the curve.

This nonlinear curve smoothing equation gives rise to an affine-invariant scale-space for images through an application of the level set method for interface propagation~\cite{set1999}. Given image $u$, the level contours of $u$, namely the sets
\begin{equation}
K_c = \left\{ \left(x,y\right)\in \mR^2 ~|~ u\left(x,y\right) = c\right\}
\end{equation}
for each $c\in \mR$, are made to evolve simultaneously according to PDE~\eqref{eq:ce}. Notice that differentiating with respect to $t$ along the $k$-level contour we obtain
\begin{equation}
\frac{d}{dt}u\left(x,y\right) = u_t + \nabla u \cdot C_t = 0
\label{d:1}
\end{equation}
where subscripts indicate partial derivatives. If we let this curve evolve according to PDE~\eqref{eq:ce}, 
\begin{equation}
C_t = \kappa^{1/3}N = -\kappa^{1/3} \frac{\nabla u}{\left\| \nabla u \right\|}
\label{d:2}
\end{equation}
where we used the fact that the gradient of a function is normal to the level curves of that function. Combining equations~\eqref{d:1} and~\eqref{d:2}, we obtain
\begin{equation}
u_t = \kappa^{1/3}\left\| \nabla u \right\|
\label{eq:aiss}
\end{equation}
Lastly, we wish to write PDE~\eqref{eq:aiss} entirely in terms of $u$ so that it may be applied directly to images without explicitly considering the level sets of the image. Hence, we require an expression for $\kappa$, the Euclidean curvature of the level contour, in terms of the image function $u$. This is well known, and can be shown to be
\begin{equation}
\kappa = \hbox{div}\left( \frac{\nabla u}{\left\| \nabla u \right\|} \right) = \frac{u_x^2u_{yy}-2u_xu_yu_{xy}+u_y^2u_{xx}}{\left(u_x^2+u_y^2\right)^{3/2}}
\end{equation}
Substitution of this into PDE~\eqref{eq:aiss} gives
\begin{equation}
u_t = \left( u_x^2u_{yy}-2u_xu_yu_{xy}+u_y^2u_{xx}\right)^{1/3}
\label{eq:pde}
\end{equation}
which is the 2D affine-invariant geometric heat equation. The affine-invariant scale space of 2D images is constructed by solving the PDE~\eqref{eq:pde} with initial data $u_0$, the original image. As with the curve evolution equation~\eqref{eq:ce}, we obtain smoothed versions of the image as we evolve in time. An example of this image smoothing is shown and compared with the linear case in Figure~\ref{fig:aiss}. This evolution is truly affine-invariant, in the sense that for two initial images $u_1\left(x\right)$ and $u_2\left(x\right)$ related by an equi-affine transformation, the scale-space images at any time $t>0$, namely $u_1\left(x,t\right)$ and $u_2\left(x\right)$, are related by the same equi-affine transformation.

In our feature detection pipeline, the affine-invariant scale-space of an image is first computed by solving PDE~\eqref{eq:pde} via finite differences. As in traditional linear methods, we sample the scale-space at six or eight discrete times, and we search for candidate feature points over each sample image. In this way, we capture details at several different scales and levels of smoothness.

These invariant flows are examples of a more general framework, developed in \cite{olv1997,olv1994}, for constructing invariant curve and surface flows under general Lie groups. Of future interest, the curve flow PDE~\eqref{eq:aiss} was extended in~\cite{olv1997} to an affine-invariant surface evolution equation
\begin{equation}
S_t = \kappa_+^{1/4} \left\| \nabla S \right\|
\end{equation}
where $S$ is the surface in 3D and $\kappa_+ = \hbox{max}\left\{ 0, \kappa \right\}$ is the (non-negative) mean curvature of $S$. Similar to above, this equation generates an affine-invariant scale-space for 3D images and provides invariant smoothing with edge preservation.

\subsection{Examples}

Once feature points are detected in both images, SURF descriptors are computed for each feature point~\cite{bay2006}. These descriptors represent smoothed versions of the local Hessian matrix about the pixel of interest, which are known to represent local shape. These vectors are not affine-invariant, and so we apply an affine region normalization procedure prior to descriptor computation~\cite{bau2000}. Points are matched between images by comparing their descriptor vectors. Different metrics can be used to assess the similarity between features. For example, the traditional Euclidean difference between the feature vectors is a simple choice, which we use here.

Once matches (correspondences) are found between the feature points, we may now align the point sets. Traditionally, an affine transformation can be fit to the correspondences by the least-squares method. However, this method considers every correspondence as equal. In reality, feature algorithms often return erroneous matches, and so we need a fitting algorithm that is robust to mismatches. One such robust estimation method is RANSAC for model fitting~\cite{fis1981,col2006}. In this algorithm, candidate affine transformation models are generated by selecting a minimal set of correspondences and computing the transformation based on only these points. Then, the number of correspondences that agree (within some threshold) with this model are counted and are called inliers. The entire process is repeated many times with different randomly-selected sets of correspondences. Ultimately, the generated transformation with the largest number of inliers is selected as the ideal model. In this manner, outliers are filtered out, since models generated using outliers will not have a large number of inliers.

\begin{figure}[h]
\centering
\subfloat[]{\includegraphics[scale=.1]{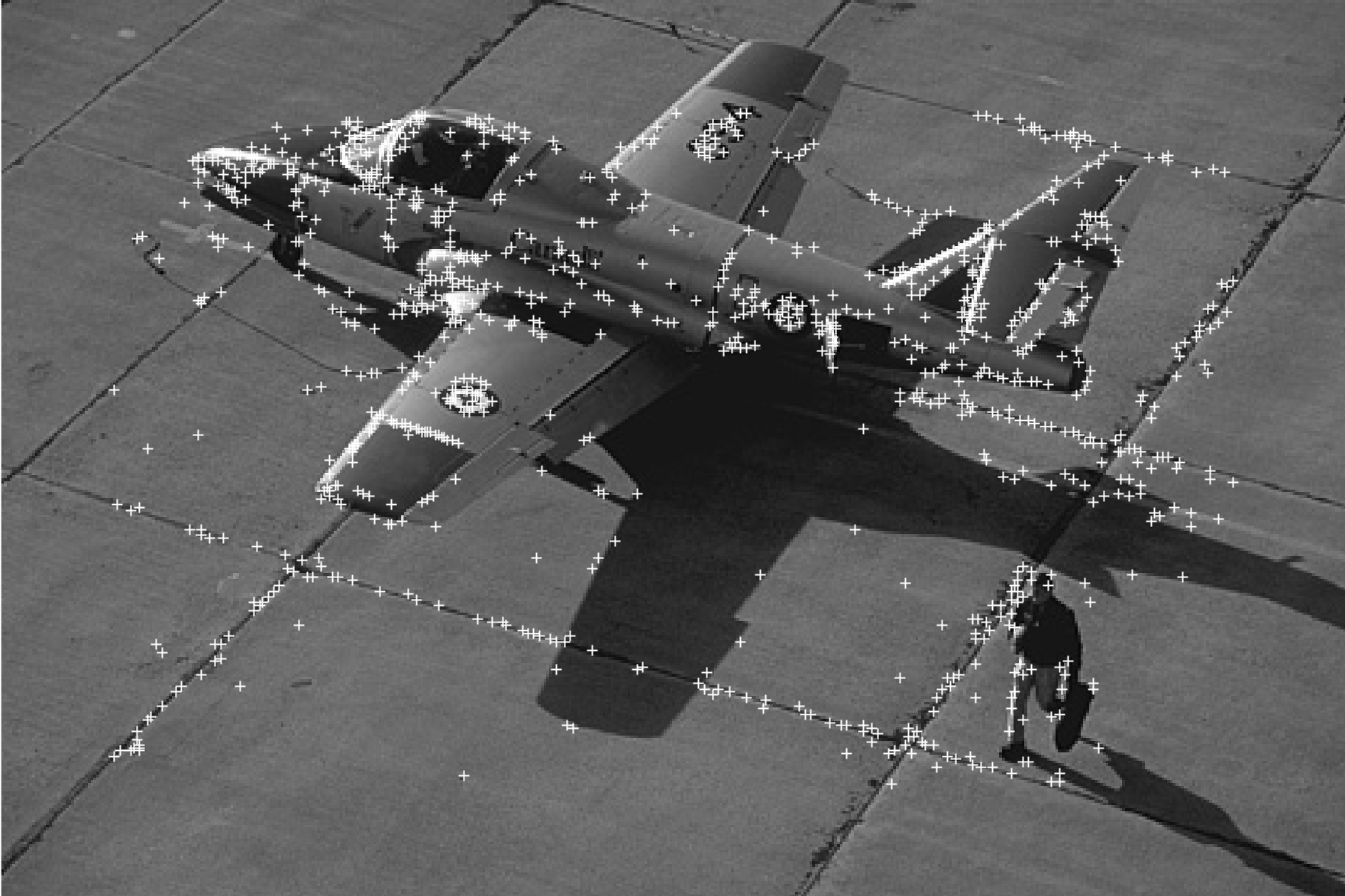}} ~
\subfloat[]{\includegraphics[scale=.1]{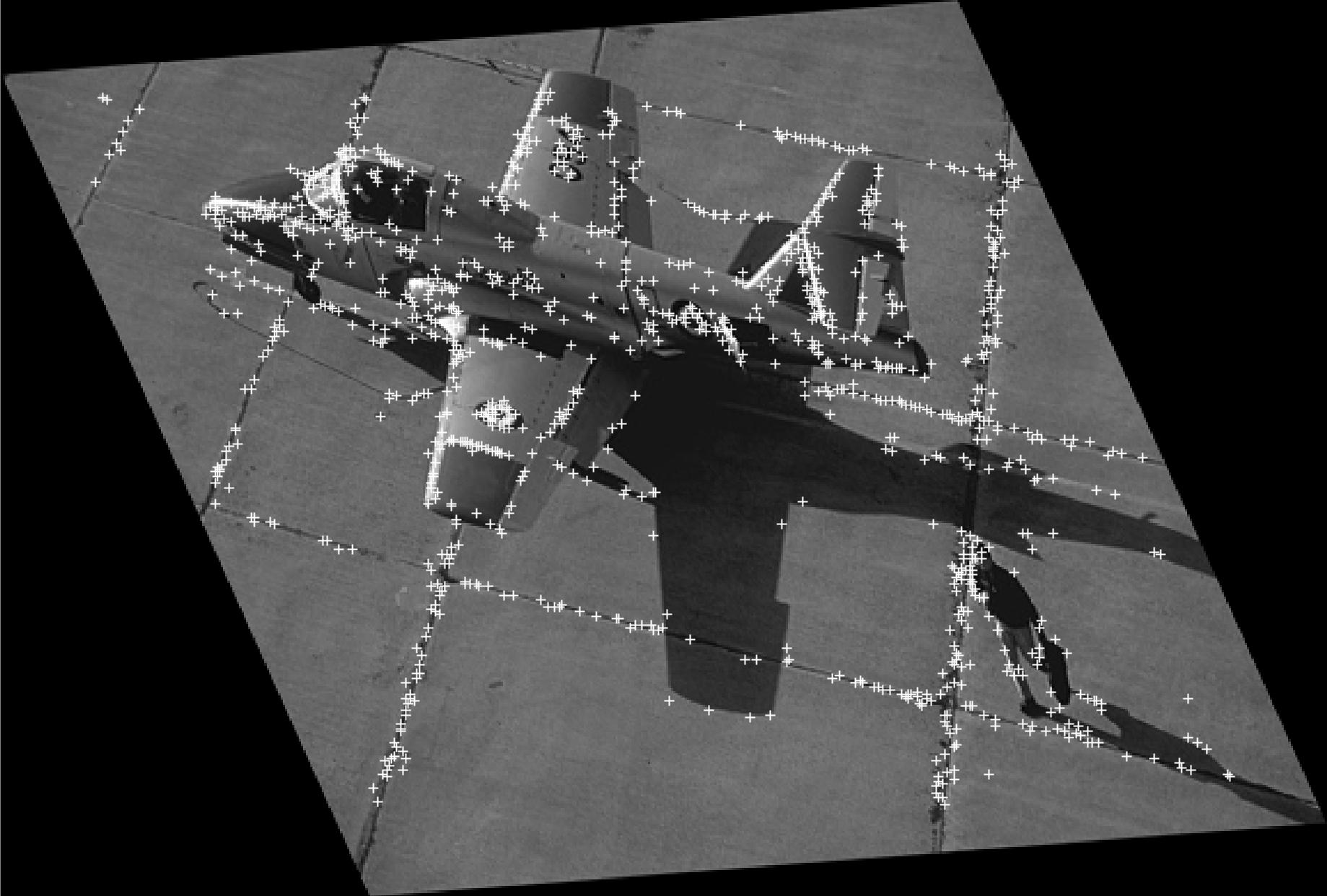}}  \\
\subfloat[]{\includegraphics[scale=.2]{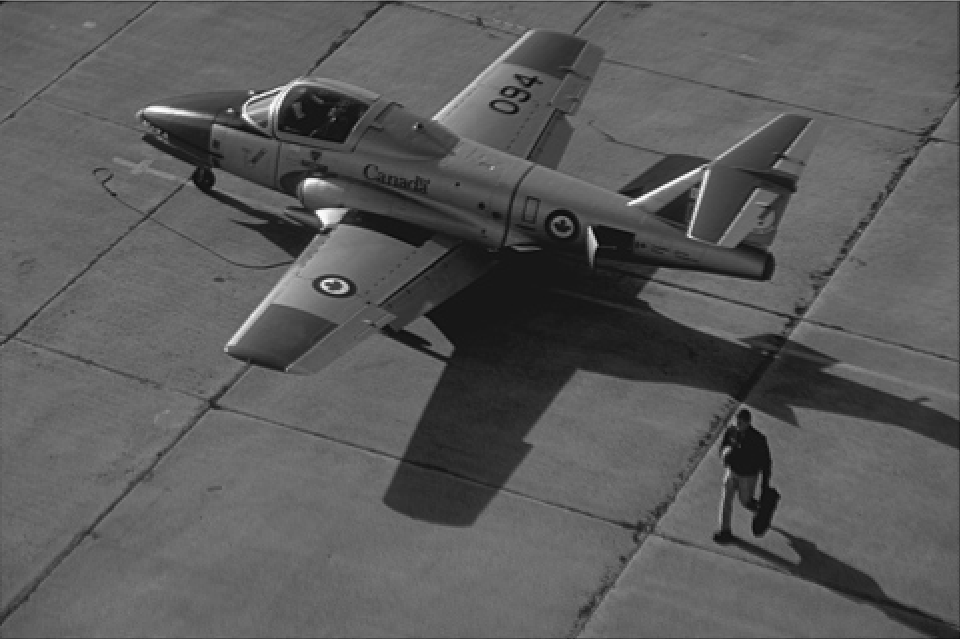}}
 \caption{Affine-invariant gradient magnitudes of the images in Figure~\ref{fig:ims}, thresholded for clarity.}
 \label{fig:ex1}
\end{figure}

\begin{figure}[h]
\centering
\subfloat[]{\includegraphics[scale=.1]{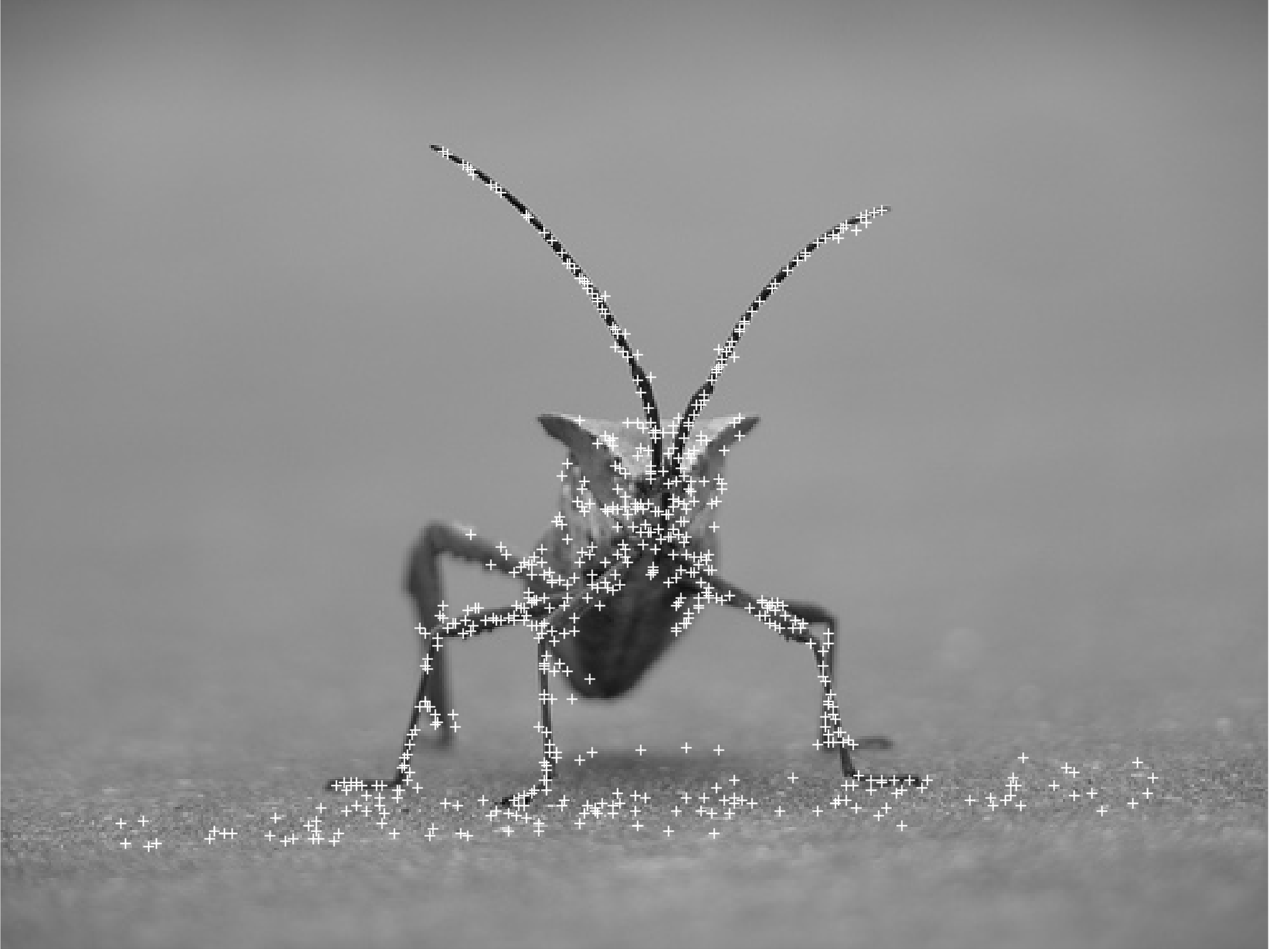}} ~~
\subfloat[]{\includegraphics[scale=.1]{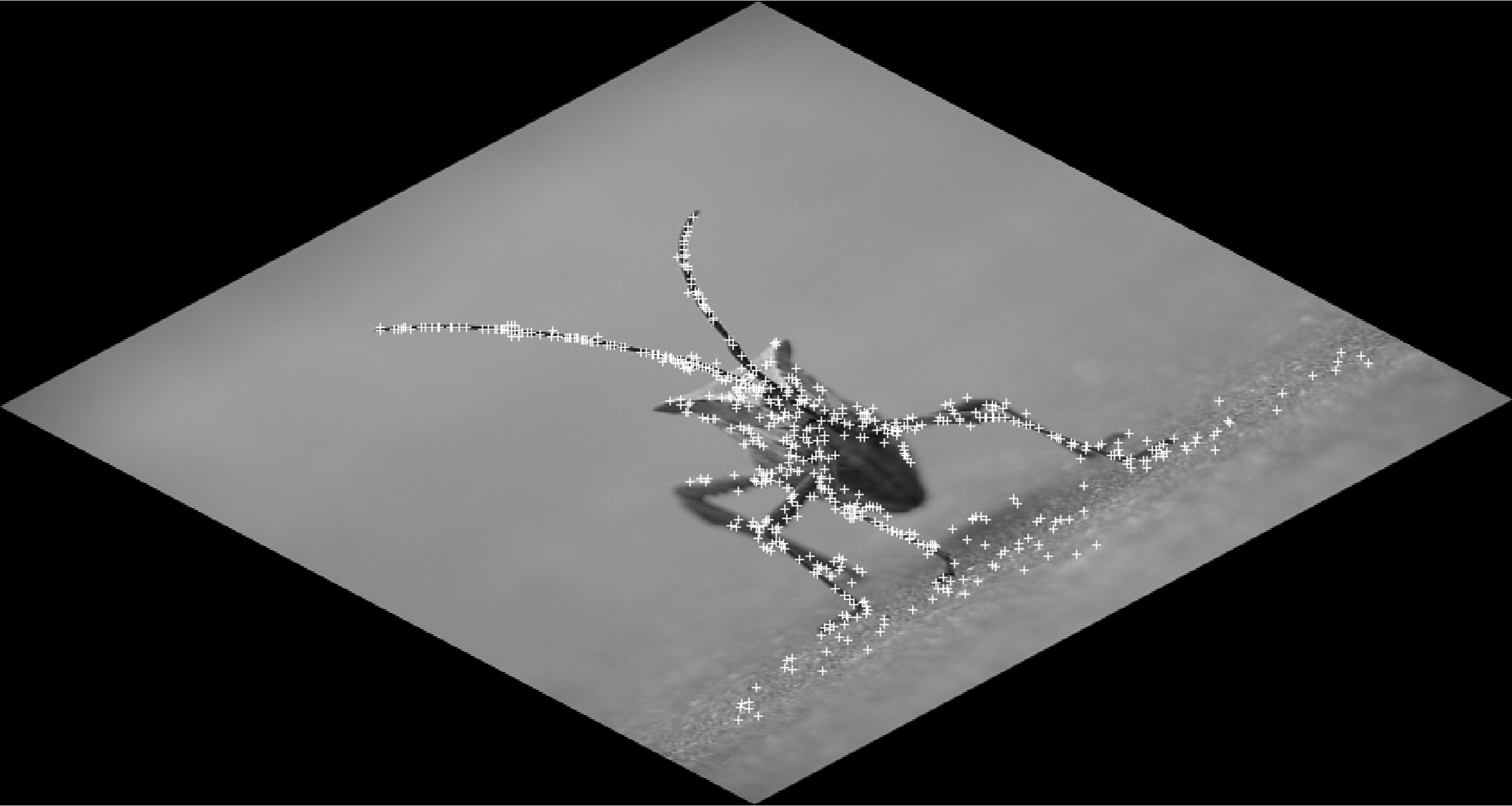}} \\
\subfloat[]{\includegraphics[scale=.2]{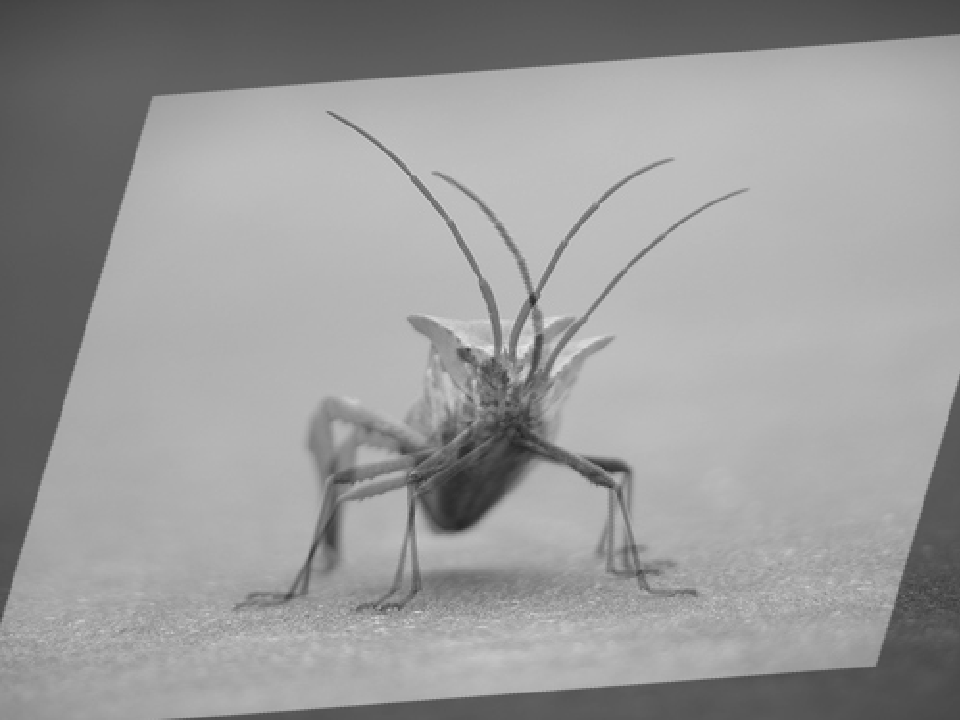}}
 \caption{Feature points and alignment of an equi-affine related image pair. The homogeneity of the background and the presence of unremarkable edges leads to an imperfect alignment. }
 \label{fig:ex3}
\end{figure}

Example alignments of affine transform image pairs are shown in Figures~\ref{fig:ex1},~\ref{fig:ex2}, and~\ref{fig:ex3}. In Figure~\ref{fig:ex1}, the computed transformation does a very good job of aligning the images; any differences between the overlaid images are difficult to notice. This image has a lot of variety, and so we expect to do well here.

The computed affine transform in Figure~\ref{fig:ex2} exhibits errors. This is likely due to the affine-invariant gradient detector exhibiting strong response along object edges. Edge points are difficult to match since they often only vary in one direction, and so could be believably matched to any other pixel along the same edge. The transform shown in Figure~\ref{fig:ex3} suffers from the same limitations. 

Another limitation of the method is that the differential invariant computations are only approximations via finite differences using traditional rectangular grids. As such, they are imperfect, and even a close inspection of Figure~\ref{fig:hex} will reveal that the responses for the two images are not exactly equal up to affine transformation. This results in inaccurate localization of feature points, and compounds the aforementioned issue with the edge points. A further investigation might pursue more accurate numerical derivative approximations, but these would slow the algorithm.

Despite these limitations, the algorithm performs well for using no prior information. The method might be used as a rough initial registration, after which a more sophisticated deformable registration might be applied. Registration could be improved by incorporating other detectors (for example Harris corners) which, though perhaps not affine-invariant, may return points which could be reliably matched.

\begin{figure}[h]
\centering
\subfloat[]{\includegraphics[scale=.2]{chairpts.png}} ~~
\subfloat[]{\includegraphics[scale=.2]{chairpts2.png}} \\
\subfloat[]{\includegraphics[scale=.2]{roughalign.png}}
 \caption{Feature points and alignment of an equi-affine related image pair. The homogeneity of the background and the presence of unremarkable edges leads to an imperfect alignment. }
 \label{fig:ex2}
\end{figure}

\section{Future Directions}
We have computed the fundamental equi-affine differential invariants for 3D image volumes. A future investigation will focus on the application of these invariants to an analogous invariant feature point detection and registration pipeline.

Moving frames have been used to design group-invariant numerical approximations, as in~\cite{cal1996,wel2007}. These methods may be applied to develop equi-affine finite difference approximations to the differential invariants, which would better serve to localize invariant feature points. 

The most interesting invariants for computer vision applications are for the 3D-to-2D projective group, since these transformations model how the real world is projected onto the image plane of a camera. Unfortunately, the projective group invariants are numerically difficult in that they involve derivatives of orders higher than two, and this is prohibitive for applications in which speed is a priority. The computation of projective invariants is being investigated. Related works connect the differential invariants of 3D curves and their 2D projections through the method of moving frames~\cite{bur2013,kog2015}.

\section{Conclusion}
In this paper, we have shown how the equivariant method of moving frames is used to compute the fundamental second-order differential invariants of the (special) affine group acting on scalar functions on $\mR^2$. These invariants were used to construct an affine-invariant feature point detector function, which was demonstrated to perform well at alignment of affine-related image pairs. The differential invariants for the equi-affine group acting on 3D image volumes are also computed, and the extension of the 2D pipeline to 3D image volumes (e.g., MRI) is an interesting future direction.

\newpage

{
\bibliographystyle{IEEEtran}
\bibliography{./refs}
}

\end{document}